\DeclareUnicodeCharacter{FB01}{fi}
\documentclass[journal]{IEEEtran}
\usepackage{amsmath}
\usepackage{epsfig}
\usepackage{graphicx}
\usepackage{amsmath}
\usepackage{amssymb}
\usepackage{multirow}
\usepackage{subfigure}
\usepackage{booktabs} \usepackage{booktabs}    
\usepackage{comment}

\def\RsetD{I\mkern-6mu R}

\usepackage{amsmath}
\newtheorem{theorem}{Theorem}

\newtheorem{proof}{Proof}
\usepackage[linesnumbered,boxed,ruled,commentsnumbered]{algorithm2e}
\usepackage[colorlinks,
            linkcolor=blue,
            anchorcolor=blue,  
            citecolor=blue,       
            ]{hyperref}

\newsavebox\CBox
\def\textBF#1{\sbox\CBox{#1}\resizebox{\wd\CBox}{\ht\CBox}{\textbf{#1}}}

\begin{document}

\title{Lifelong Mixture of Variational Autoencoders}

\author{\IEEEauthorblockN{Fei Ye and Adrian G. Bors, {\em Senior Member, IEEE}}

\IEEEauthorblockA{Department of Computer Science, University of York, York YO10 5GH, UK \\
E-mail: fy689@york.ac.uk,
adrian.bors@york.ac.uk }
}


\maketitle

\begin{abstract}
 In this paper, we propose an end-to-end lifelong learning mixture of experts. Each expert is implemented by a Variational Autoencoder (VAE). The experts in the mixture system are jointly trained by maximizing a mixture of individual component evidence lower bounds (MELBO) on the log-likelihood of the given training samples. The mixing coefﬁcients in the mixture model, control the contributions of each expert in the global representation. These are sampled from a Dirichlet distribution whose parameters are determined through non-parametric estimation during the lifelong learning. The model can learn new tasks fast when these are similar to those previously learnt. The proposed Lifelong mixture of VAE (L-MVAE) expands its architecture with new components when learning a completely new task. After the training, our model can automatically determine the relevant expert to be used when fed with new data samples. This mechanism benefits both the memory efficiency and the required computational cost as only one expert is used during the inference.  The L-MVAE inference model is able to perform interpolations in the joint latent space across the data domains associated with different tasks and is shown to be efficient for disentangled learning representation. The code is available~: \url{https://github.com/dtuzi123/LifelongMixtureVAEs} 
\end{abstract}

\begin{IEEEkeywords}
Lifelong learning, Mixture of Variational Autoencoders, Multi-task learning, Mixture of Evidence Lower Bounds, Disentangled representations.
\end{IEEEkeywords}

\IEEEpeerreviewmaketitle

\section{Introduction}
\label{sec:intro}
Deep learning models suffer from catastrophic forgetting \cite{LifeLong_review}  when trained on multiple databases in a sequential manner. A deep learning model quickly forgets the characteristics of the previously learned experiences while adjusting to learning new information. The ability of artificial learning systems of continuously acquiring, preserving and transferring skills and knowledge throughout their lifespan is called lifelong learning \cite{LifeLong_review}. Existing approaches would either use dynamic architectures, adopt regularization during training, or employ generative replay mechanisms. Dynamic architecture approaches  \cite{Adanet,ProgressiveNN,Error_driven,OnlineLearning,LearnAdd} would increase the network capacity by adding new layers and processing units in order to adapt the network's architecture to acquiring new information. However, such approaches would require a specific architecture design while their parameters would increase progressively with the number of tasks. Regularization approaches \cite{Distilling_nets,LessForgetting,EWC,Lwf,LifeLong_combination} aim to impose a penalty when updating the network' parameters in order to preserve the knowledge associated with previously learned tasks. In practice, these approaches suffer from performance degradation when learning a series of tasks where the datasets are entirely different from the previously learned ones. Memory-based methods use a buffer in order to upload previously learned data samples \cite{TinyLifelong,GradientLifelong}, or utilize powerful generative networks such as a Variational Autoencoders (VAEs) \cite{Generative_replay,GenerativeLifelong,Lifelong_VAE,LifelongUnsupervisedVAE} or Generative Adversarial Networks (GANs) \cite{LGAN,MemoryGAN}  as memory-based replay networks that reproduces and generates data which is consistent with what has seen and learned before. These approaches would need additional memory storage space for recording parameters of the generated data while their performance on the previously learned tasks is heavily dependent on the generator's ability to realistically replicate data.

Promising results have been achieved on prediction tasks \cite{LearnAdd,Distilling_nets,ImprovedAGEM,CalibratingCNN,ThreeScenarios_Lifelong,Continual_Learning}. However, these methods do not capture the underlying structure behind the data, which prevents them from being applied in a wide range of applications. There are very few attempts addressing representation learning under the lifelong setting \cite{GenerativeLifelong,Lifelong_VAE}. The performance of these methods degrades significantly when engaging in the lifelong training with datasets containing complex images or on a long sequence of tasks. The reason is that these approaches require to retrain their generators on artificially generated data. Meanwhile, the performance loss on each dataset is accumulated during the lifelong learning of a sequence of several tasks. To address this problem, we propose a probabilistic mixture of experts model, where each expert infers a probabilistic representation of a given task. A Dirichlet sampling process defines the likelihood of a certain expert to be activated when presented with a new task.  

This paper has the following contributions~: 
\begin{itemize}
	\item  A novel mixture learning model, called Lifelong Mixture of VAEs (L-MVAE). Instead of capturing different characteristics of a database as in other mixture models  \cite{VMVAE,MSVAE,Cluster_VAE,Multimodal_generative}, the proposed mixture model enables to automatically embed the knowledge associated with each database into a distinct latent space modelled by one of the mixture's experts during the lifelong learning. 
	\item A training algorithm based on the maximization of the mixture of individual component evidence lower bounds (MELBO)
    \item A mixing-coefficient sampling process is introduced in order to activate or drop out experts in L-MVAE. Besides defining an adaptive architecture, this procedure accelerates the learning process of new tasks while overcoming the forgetting of the previously learned tasks.
\end{itemize}
The remainder of the paper contains a detailed overview of the existing state of the art in Section~\ref{RelWorks}, while the proposed L-MVAE model is discussed in Section~\ref{LifeLongMVAE}. In Section~\ref{TheoAnal} we discuss the theory behind the proposed L-MAE model and in Section~\ref{ExtendLearning} we explain how the proposed methodology can be used in unsupervised, supervised and semi-supervised learning applications. The expansion mechanism for the model's architecture is presented in Section~\ref{MEM}. The experimental results are analyzed in Section~\ref{Exper} while the conclusions are drawn in Section~\ref{Con}.

\section{Related research studies}
\label{RelWorks}

A variational autoencoder (VAE) \cite{VAE} is made up of two networks, an encoder and a decoder. Given a data set, the encoder extracts a latent vector ${\bf z}$, and the decoder aims to reconstruct the given data from the latent vectors.
A number of research works have been developed for capturing meaningful and disentangled data representations by using the VAE framework \cite{baeVAE,DisentanglingByFactorising,VAETCE,UnVAE,DisentanglingDistanglement}. These approaches show promising results on achieving disentanglement between latent variables as well as interpretable visual results, where specific properties of the scene can be manipulated through changing the relevant latent variables. However, these models work well only on data samples drawn from a single domain, corresponding to a specific database used for training. When they are re-trained on a different database, their parameters are updated and then they fail to perform on the tasks learned previously. This happens because they do not have appropriate objective functions to deal with catastrophic forgetting \cite{EWC,Catastrophic,OnlineStructuredLaplace}. 

Recently, there have been some attempts to learn cross-domain representations under the lifelong learning by introducing an environment-dependent mask that specifies a subset of generative factors \cite{Lifelong_VAE}, by proposing a Teacher-Student lifelong learning framework  \cite{GenerativeLifelong,LTS} or a hybrid model \cite{LatentVAEGAN} of Generative Adversarial Nets (GANs) \cite{GAN} and VAE. The models proposed in \cite{GenerativeLifelong,Lifelong_VAE,LatentVAEGAN} are based on Generative Replay Mechanisms (GRM) aiming to overcome forgetting. However, these methods suffer from poor performance when considering complex data. 

Aljundi {\em et al.} \cite{Expertgate} proposed a lifelong learning system named the Expert Gate model, where new experts are added to a network of experts. The most relevant expert from the given set is chosen during the testing stage, according to the reconstruction error of the data. However, this may not necessarily correspond to the best log-likelihood estimate for the data. Moreover, the Expert Gate model was used only for supervised classification tasks. 

Regularization based approaches alleviate catastrophic forgetting by adding an auxiliary term that penalizes changes in the weights when the model is trained on a new task  \cite{LearnAdd,Distilling_nets,LessForgetting,EWC,Lwf,LifeLong_combination,OnlineStructuredLaplace,BoostingTransfer,VCL,CL_Bayesian} or store past samples to regulate the optimization \cite{ImprovedAGEM,AGEM}. However, regularization based approaches
have huge computation requirements when the number of tasks increases \cite{GradientEpisodic}.

In another direction of research, mixtures of VAEs have been employed for continuous learning \cite{VMVAE,MSVAE,Cluster_VAE,Multimodal_generative}. These models are able to capture underlying complex structures behind data and therefore perform well on many down-stream tasks including clustering and semi-supervised classification. However, these mixture models would only capture characteristics of a single database which had been split into batches of data, and tend to forget previously learned data characteristics when attempting to learn a sequence of distinct tasks. In contrast to the above mentioned methods, our model is able to capture underlying generative latent variable representations across multiple data domains during the lifelong learning.

\section{The Lifelong Mixture of VAEs}
\label{LifeLongMVAE}

\subsection{Problem formulation}

In this paper we consider a model made up of a mixture of networks \cite{MixVAE} which is able to deal with three different learning scenarios: supervised, semi-supervised and unsupervised, under the lifelong learning setting. Let us consider a sequence of tasks and denote 
${\cal D}^{(k)} = \{ {\bf x}_i^{(k)},{\bf y}_i^{(k)}\}_i^{N_k}$ as a dataset characterizing the $k-th$ task, where 
${\bf x}_i^{(k)} \in {\cal X}^{(k)}$ is the source domain and 
${\bf y}_i^{(k)} \in {\cal Y}^{(k)}$ is the target domain which is usually defined by class labels, while each domain $\{ {\cal D}^{(i)} | i=1,\ldots,K \}$ is associated to a given task.
 We aim to learn a model which not only generates or reconstructs data but which can also generate meaningful representations useful for various tasks during a lifelong learning process.

\subsection{Mixture objective function}

Traditional mixture models \cite{VGMM,SparseMulGMR} normally capture different characteristics of a dataset by learning several latent variable vectors, with distinct sets of variables associated to each mixture' component. In this paper, we implement each expert by using a generative latent variable model $p_\theta ({\bf x}, {\bf z}) = p_\theta ( {\bf x}|{\bf z})p({\bf z})$, where
${\bf z} \in \RsetD^d$ is the latent variable and $\theta$ 
represents the decoder's parameters, as in VAEs \cite{VAE}. The learning goal of the generative model is to maximize the log-likelihood of the data distribution, which is actually a difficult problem due to the intractability of the marginal distribution 
$p ( {\bf x}) = \int p_\theta ( {\bf x}|{\bf z})p ({\bf z})d{\bf z}$, requiring access to all latent variables. Instead, we optimize the evidence lower bound (ELBO) on the data log-likelihood, \cite{VAE}~:
\begin{equation}
\begin{aligned}
 \log p ({\bf x}) &\ge  {\mathbb{E}_{{\bf z} \sim {q_{\varepsilon} } ({\bf z}|{\bf x})}[ \log p_{\theta} ({\bf x}|{\bf z})}] - D_{KL} [ q_{\varepsilon} ({\bf z}|{\bf x} )||p ({\bf z}) ] \\&={\mathcal L}_{\rm{VAE},\theta ,\varepsilon } ({\bf x}), 
\end{aligned}
\label{DefineL_VAE}
\end{equation}
where $q_{\varepsilon} ({\bf z}|{\bf x})$ is called the variational distribution, and $\varepsilon$ represents the parameters of the encoder. We use the Gaussian distribution for both the prior $p({\bf z})$ as well as for the variational distribution $q_\varsigma ({\bf z}|{\bf x})$. The latent variable ${\bf z}$ is sampled using the reparametrization trick \cite{VAE}, ${\bf z}_i = {\bf u}_i + \delta  \otimes \sigma_i$, where ${\bf u}_i$ and $\sigma_i$ are inferred by the encoder, and $\delta$ is sampled from ${\cal N}(0,{\bf I})$.  $p_{\theta} ( {\bf x}|{\bf z})$ is implemented by a decoder with trainable parameters $\theta$, receiving the latent variables ${\bf z}$ and producing data reconstructions ${\bf x}'$.

When considering that we have $K$ experts in the mixture model,  we introduce the loss function as the Mixture of individual ELBOs (MELBO)
${\mathcal L}_{VAE}^i({\bf x})$, defined through (\ref{DefineL_VAE})~:
\begin{equation}
\begin{aligned}
{\mathcal L}_{L-MVAE} ({\bf x}) =\frac{\sum\limits_{i = 1}^K w_i{\mathcal L}_{VAE}^i({\bf x})}{\sum\limits_{i = 1}^K w_i}\,,
\end{aligned}
\label{MELBO}
\end{equation}
where $w_i$ is the mixing coefﬁcient, which controls the significance of the $i$-th expert. We model all mixing coefﬁcients by using a Dirichlet distribution $\{w_1,\dots,w _K\} \sim{\mathop{\rm Dir}\nolimits} ({\bf a})$, of parameters ${\bf a} = \{a_1,\dots,a_K\}$. In the following we describe the mechanism for selecting appropriate L-MVAE components during the training.

\subsection{The selection of L-MVAE mixture's components during training}
\label{SelectExpert}

Certain research studies \cite{VMVAE,MSVAE} have considered equal contributions for the components of deep learning mixture systems. However, in this paper we consider that each mixture component is specialized for a specific task. The selection of a specific mixture component is performed through the mixing weights $w_i$, $i=1,\ldots,K$. We assume that the weighting probability for each mixture's component is drawn from a Multinomial distribution, such as the Bernoulli distribution, defined by a Dirichlet prior. 

\textBF{Assignment vector. } In the following, we introduce an assignment vector ${\bf c}$, with each of its entries $c_i \in \{0,1\}$, $i=1,\ldots,K$, representing the probability of including or not the $i$-th expert in the mixture. $c_i$ is sampled from as Bernoulli distribution. Before starting the training, we set all entries as $c_i=0$, $i=1,\ldots,K$. The assignment probability for each mixing component is calculated considering the sample log-likelihood of each expert after learning each task, as~:
\begin{equation}
\begin{aligned}
p(c_j) = 1- \frac{\exp ( - \mathcal{L}_{VAE}^j({\bf x}_b)) +  u\;c'_j}{\sum\limits_{i = 1}^K \left( \exp ( -\mathcal{L}_{VAE}^i({\bf x}_b)) + u\;c'_i \right) } \,,
\label{selection}
\end{aligned}
\end{equation}
where ${\bf x}_b$ is sampled from the given data batch, drawn from the database corresponding to the current task learning. $c'_j$ denotes the assignment variable for $j$-th expert and represents the value resulted when learning the previous task before evaluating Eq.~(\ref{selection}). $u \, c'_j$ is used to ensure that $p(c_j)$ is outside the range of possible values for $c'_j=1$, when evaluating Eq. (\ref{selection}), and therefore we consider $u$ as a large value. Then we find the maximum probability for a mixing component~:
\begin{equation}
p(c_{j^*}) = \max (p(c_1),\ldots,p(c_K))\,,
\label{findOptimal}
\end{equation}
where $j^*$ represents the index of the selected VAE component according to the parameters learnt during the previous tasks.
We then normalize the other assignment variables, except for $j^*$:
\begin{equation}
\begin{aligned}
p(c_i) = \begin{cases}
1& c_i^\prime  = 1\\
0& c_i^\prime  = 0
\end{cases}
, \; i = 1,2,\ldots, K, \; i \ne j^*
\,.
\label{recover}
\end{aligned}
\end{equation}
Since $c_i^\prime$ is an assignment corresponding to the learning process of the previous task, before evaluating Eq.~(\ref{selection}), in order to determine the dropout status of $i$-th expert during the current task learning, we use Eq.~(\ref{recover}) to recover the dropout status of all experts except for $j^*$-th expert which is actually dropped out from the future training because it is going to be used for recording and reproducing the information associated with the current task being learnt. When learning the first task, all mixture's components will be trained and then when learning the second task, only $K-1$ components are trained, while one component is no longer trained because it is considered as a depository of the information associated with the first task. This component will consequently be used to generate information consistent with the probabilistic representation associated with the first task. This process is continued and for the last task at least one VAE is available for training. The number of mixing components $K$ considered initially should be larger or at least equal to the number of tasks assumed to be learned during the lifelong learning process. In Section~\ref{MEM} we describe a mechanism for expanding the mixture.

\textBF{The sampling of mixing weights.} Suppose that L-MVAE finished learning the $t$-th task. We collect several batches of samples $\{ {\bf x}_i,\dots, {\bf x}_N \}$ from the $(t+1)$-th task, where each ${\bf x}_i$ represents the $i$-th batch of samples, which are used to evaluate the assignment vector ${\bf c}$ by using Eq.~\eqref{selection}. We calculate the average probability $p(c_j) = \sum\nolimits_{i = 1}^N p(c_j^i)/N$, where each $p(c_j^i)$ represents the probability for the assignment of ${\bf x}_i$. Then we find $p(c_{j*})$ by using Eq.~\eqref{findOptimal} and we recover the previous assignments except for $c_{j*}$ by using Eq.~\eqref{recover}. 
The Dirichlet parameters are calculated in order to fix the mixture components containing the information corresponding to the previously learnt tasks while making the other mixture components available for training with the future tasks. For the mixing components that have been used for learning the previous tasks, we consider 
\begin{equation}
\begin{aligned}
&a_i = \begin{cases}
e & c_i = 1\\
\frac{1-e*K'}{K - K'} & c_i = 0\,, i=1,\dots,K'
\end{cases} 
\label{selection_a}
\end{aligned}
\end{equation}
where $e$ is a very small positive value. For $i=1,\ldots,K'$, where $K'$ represents the number of tasks learnt so far out of a total of $K$ given tasks, during the lifelong learning. A small value for the Dirichlet parameters implies that the corresponding mixture components are no longer trained. The mixing weights $w_1,\dots,w_K$ are sampled from a Dirichlet distribution with parameters $a_1,\dots,a_K$. We then train the mixture model with $w_1,\dots,w_K$ by using Eq.~\eqref{MELBO} at the $(t+1)$-th task learning.

\begin{figure*}[htbp]
	\centering
	\includegraphics[scale=0.31]{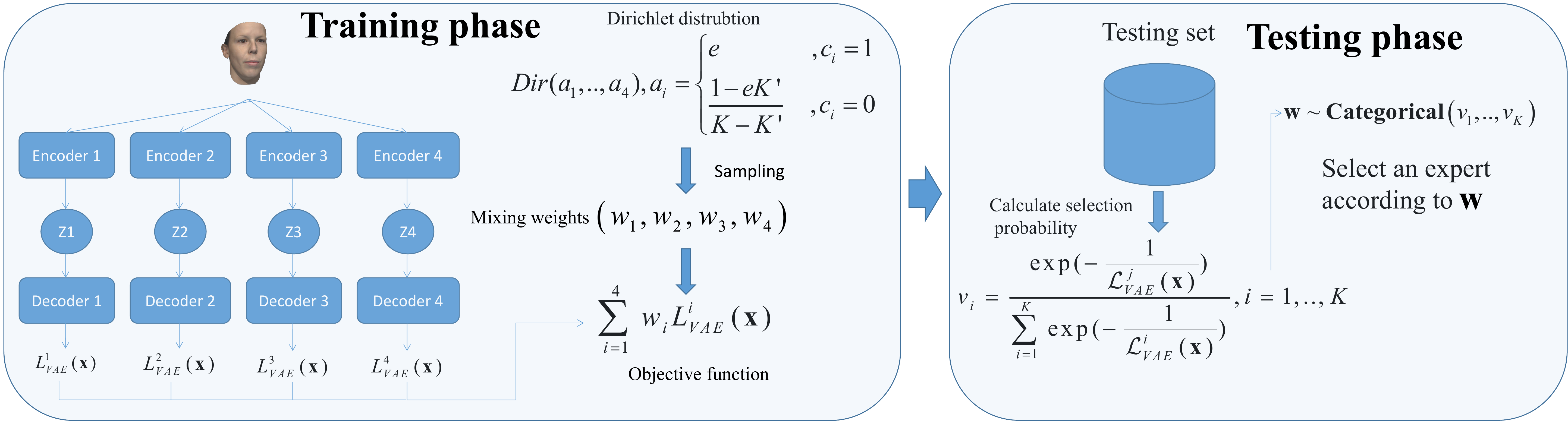}
	\caption{The structure of the proposed Lifelong Mixture of VAEs learning system with $K=4$ components. Each expert has an independent inference and generation process and therefore we can calculate $\mathcal{L}^i_{VAE}({\bf x})$, the ELBO for each expert. Each $c_i$ represents the probability of the assignment for the $i$-th component, which is used to determine $a_i$ by using Eq.~\eqref{selection_a}. The mixing weights $\{w_1,\dots,w_4 \}$ are sampled from the Dirichlet distribution and used in Eq.~(\ref{MELBO}). During the testing phase, for given data samples we select an appropriate mixture component to be used.} 
	\label{Mixture_structure}
\end{figure*}

\textBF{Testing phase.} Suppose that after the lifelong learning process, we have trained $K$ components. In the testing phase, we perform a selection of a single component to be used for the given data samples. We firstly calculate the selection probability $\{v_1,\dots,v_K\}$ by calculating the log-likelihood of the data sample for each component~:
\begin{equation}
\begin{aligned}
v_j = \frac{\exp \left( - \frac{1}{{\cal L}_{VAE}^j({\bf x})}\right)}{\sum\limits_{i = 1}^K \exp  \left( - \frac{1}{{\cal L}_{VAE}^i({\bf x})}\right)}, \;\; j = 1,\ldots,K
\label{selection_component_evaluation}\,.
\end{aligned}
\end{equation}
Then we select a component by sampling the mixing weight vector ${\bf w}$ from Categorical distribution $Cat ( v_1,\ldots,v_K )$.

The structure of the proposed L-MVAE model is shown in Fig.~\ref{Mixture_structure}. In the next section we evaluate the convergence properties of L-MVAE model during the lifelong learning.

\section{Theoretical analysis of L-MVAE}
\label{TheoAnal}

In this section, we evaluate the convergence properties of the proposed L-MVAE model during the lifelong learning. We evaluate the evolution of the objective function ${\mathcal L}_{L-MVAE} ({\bf x})$ during the training and define a lower bound on the data's log-likelihood. We also show how L-MVAE model infers across several tasks during the lifelong learning.

{\em Definition.} Let us define the following function~:
\begin{equation}
\mathcal{L}_{EMIX}({\bf x}) = \sum\limits_{i=1}^K w_i \exp(\mathcal{L}_{VAE}^i({\bf x}))
\end{equation}
where $\mathcal{L}_{VAE}^i({\bf x})$  is defined for the $i$-th mixture component by considering the objective function (\ref{DefineL_VAE}) and where we consider $\sum_{i = 1}^K w_i=1$. We also define the likelihood function for the mixture model, denoted as $\mathcal{L}_{L-Log}({\bf x})$.

{\em Lemma.} By considering  $\mathcal{L}_{L-Log}({\bf x})$, defining the likelihood function 
\begin{equation}
\mathcal{L}_{L-Log}({\bf x}) = \sum\nolimits_{i = 1}^K w_i\int p_{\theta_i} ( {\bf x}|{\bf z}) p_{\theta_i}( {\bf z})d{\bf z},
\end{equation}
and the previous {\em Definition}, we have~:
\begin{equation}
\log \mathcal{L}_{L-Log}({\bf x}) > \log \mathcal{L}_{EMIX}({\bf x}).
\end{equation}

{\em Proof.} 
After considering the latent variables ${\bf z}$  for each VAE component, the marginal log-likelihood of the mixture is given by:
\begin{equation}
\begin{aligned}
\log \mathcal{L}_{L-Log}({\bf x}) & = \log \left(\int \sum\nolimits_{i= 1}^K w_i p_{\theta_i} ({\bf x},{\bf z}) d{\bf z}\right) \\&= \log \left(\sum\nolimits_{i= 1}^K w_i \int p_{\theta_i}({\bf x}|{\bf z}){p_{{\theta _i}}}({\bf z}) d {\bf z}\right)\,,
\end{aligned}
\label{MVAE}
\end{equation}

We know that $\log p_{\theta_i}({\bf x}) =\log \int p_{\theta_i}({\bf x}|{\bf z})p_{\theta_i}({\bf z}) d{\bf z}$ is bounded by the local ELBO objective function ${\mathcal L}_{VAE,\theta_i,\varepsilon_i}^i ({\bf x})$,
according to (\ref{DefineL_VAE}), and we have 
\begin{equation}
\int p_{\theta_i}({\bf x}|{\bf z})p_{\theta_i}({\bf z})d{\bf z} \ge \exp ({{\mathcal L}_{VAE,\theta_i,\varepsilon_i}^i({\bf x})})\,,
\end{equation}
where $\theta_i$ represents the parameters for the $i$-th mixture component.

Since the $\log$ function is a monotone increasing function, then we have:
\begin{equation}
\begin{aligned}
& \log \left(\sum\nolimits_{i=1}^K w_i \int p_{\theta_i}({\bf x}|{\bf z})p_{\theta_i}({\bf z}) d{\bf z}\right) \ge  \\& \log \left(\sum\nolimits_{i=1}^k w_i \exp ({\mathbb E}_{q_{\varepsilon_i} ( {\bf z}|{\bf x})} 
[ \log p_{\theta_i}({\bf x}|{\bf z}) - 
\log q_{\varepsilon_i} ( {\bf z}|{\bf x}) ])\right)\,,
\end{aligned}
\end{equation}
which proves the {\em Lemma}.

\begin{theorem}
Optimizing the mixture's objective function, $\mathcal{L}_{L-MVAE} ({\bf x})$, corresponds to finding a lower bound on the data log-likelihood, $\log \mathcal{L}_{L-Log} ({\bf x})$.
\end{theorem}

\begin{proof}
From the {\em Lemma}, we have~:
\begin{equation}
\begin{aligned}
\log \mathcal{L}_{L-Log} ({\bf x}) &= \log \left(\sum\nolimits_{i = 1}^K w_i \int p_{\theta_i} ({\bf x}|{\bf z}) p_{\theta_i}({\bf z})d{\bf z}\right) \\&\ge \log \left(\sum\nolimits_{i = 1}^K w_i 
\exp ({\mathcal L}_{VAE,\theta_i,\varepsilon_i}^i ({\bf x}))\right)\,,
\end{aligned}
\label{Theo1}
\end{equation}

When we optimize the mixture's objective function $\mathcal{L}_{L-MVAE}({\bf x})$, the loss $\mathcal{L}_{VAE,\theta_i,\varepsilon_i}^i({\bf x})$ corresponding to each component is increased. Then, the right hand side from (\ref{Theo1}) will be increased to approximate $\log {\mathcal{L}}_{L-Log}(\bf x)$, given that the $\log \bf{u}$ is monotonically increasing for ${\bf u} \in [0, + \infty )$. However, any increase has an upper limit in $\mathcal{L}_{L-Log} ({\bf x})$, according to (\ref{Theo1}) $\Box$
\end{proof}

\begin{theorem}
Let us define 
$\log \mathcal{L}^{\star}_{L-Log}({\bf x})$ as the log-likelihood of the objective function ${\mathcal{L}_{L-MVAE}}(\bf x)$. Then we have $\log \mathcal{L}^{\star}_{L-Log}({\bf x})  \le  \max\{ \log p_{\theta_i}({\bf x})\},$ $\forall i \in \{1,\ldots,K\}$ during the inference, where $\log p_{\theta_i}({\bf x})$ represents the log-likelihood of a single VAE, characterized by parameters $\theta_i$.
\end{theorem}

\begin{proof}
Estimating the log-likelihood $\log \mathcal{L}^{\star}_{L-Log}({\bf x})$ during the inference is intractable because the generation process of the mixture model involves an implicit component selection procedure. By considering (\ref{MELBO}), the log-likelihood $\log \mathcal{L}^{\star}_{L-Log}({\bf x})$ is given by~:
\begin{equation}
\begin{aligned}
\log \mathcal{L}^{\star}_{L-Log}({\bf x}) = \log \left(\sum\limits_{i = 1}^K w_i{\cal L}_{VAE,{\theta _i},{\varepsilon _i}}^i({\bf x}) \right)\,,
\end{aligned}
\end{equation}
where $\sum\nolimits_{i = 1}^K w_i  = 1$, where the mixing parameters $w_i$ are sampled from $Cat({\bf \tau} )$, ${\bf \tau}= (\tau_1 \, \tau_2 \ldots \tau_K)^T$ where $\tau_i = \log p_{\theta_i}({\bf x})/\sum\nolimits_{j = 1}^K \log p_{\theta_j}({\bf x})$. The marginal log-likelihood $\log p_{\theta_i}({\bf x})$ for each VAE component is given by its approximation ${\cal L}_{VAE,\theta_i,\varepsilon_i}^i({\bf x})$. The proposed model selects only the most suitable expert VAE, indexed as $h$, which has the highest likelihood for the given data samples used during the training~:
\begin{equation}
\begin{aligned}
&\log \mathcal{L}^{\star}_{L-Log} ({\bf x}) \le \mathcal{L}^h_{{VAE}_{{\theta _h},{\varepsilon _h}}}({\bf x}),\\
& \log p_{\theta_h} ({\bf x}) \ge \mathcal{L}^h_{{VAE}_{{\theta _i},{\varepsilon _i}}}({\bf x}),
\end{aligned}
\end{equation}
where $\log p_{\theta_h}({\bf x}) = \max\{\log p_{\theta_i}({\bf x}) \}$, $i=1,\dots,K$ $\Box$ 
\end{proof}

This shows that during the testing stage, we can evaluate the data's log-likelihood, by using the proposed L-MVAE model. Unlike in the approach from \cite{Expertgate}, the proposed mixture system not only that can perform generation tasks but it also learns meaningful data representations across the domains assimilated during the lifelong learning process.

\section{Defining the Lifelong MVAE for supervised, semi-supervised and unsupervised learning}
\label{ExtendLearning}

In this section, we extend the mixture model for being used under various types of learning paradigms, such as~: unsupervised, supervised, and semi-supervised.

\noindent \textbf{Unsupervised disentangled representation learning.} In order to encourage the latent representations to capture meaningful data variations under unsupervised learning assumptions, we extend the disentangled representation learning approach from \cite{UnVAE}, which was built on a similar concept to the $\beta$-VAE \cite{baeVAE}, for modelling disentangled representations in single VAE models. We extend \cite{UnVAE} to be used for the mixture objective function by replacing Eq.~(\ref{MELBO}) with the following loss function:
\begin{equation}
\begin{aligned}
L_{L-MVAE,\tilde{\varepsilon}, \tilde{\theta}}^{US}({\bf x}) = \sum\limits_{i = 1}^K w_i \left( - \gamma |{D_{KL}}(q_{{\varepsilon _i}} ({{\bf{z}}_i}|{\bf{x}})||p({{\bf{z}}_i})) - C| \right. \\
\left. + {\mathbb{E}_{{{\bf{z}}_i} \sim {q_{\tilde{\varepsilon _i}}}({{\bf{z}}_i}|{\bf{x}})}}\log {p_{\tilde{\theta _i}}}({\bf{x}}|{{\bf{z}}_i})\right) 
\label{UnDisentangled_loss}
\end{aligned}
\end{equation}
where $\sum_{i = 1}^K w_i$. The former term represents the Kullback-Leibler (KL) divergence associated with the output of each VAE decoder, by considering the disentanglement among the latent space variables, weighted by $w_i$, while the latter term is associated with the log-likelihood of data reconstruction by each mixture's encoder. The parameters associated with the disentanglement are set similarly to those from \cite{baeVAE}: $C$ is linearly increasing during the training, starting from a low value, while $\gamma$ defines the contribution of this modified KL term to the objective function. $\tilde{\varepsilon}=\{ \tilde{\varepsilon}_i | i=1,\ldots,K  \}$ and $\tilde{\theta} = \{ \tilde{\theta}_i | i=1,\ldots,K \}$ represent the parameters for all encoders and decoders of the mixture and of the individual components, respectively.

\noindent \textbf{Lifelong supervised learning.} We consider that the given data $\{ {\cal X} | {\bf x}_i\in {\cal X}, i=1,\ldots,N\}$ is labelled $\{ {\cal Y} | {\bf y}_i\in {\cal Y}, i=1,\ldots,N\}$, within a supervised learning framework. When considering a single VAE component we define a latent generative variable model $p_\theta ({\bf x},{\bf z},{\bf d}) = p_\theta ({\bf x}|{\bf z},{\bf d})p({\bf z},{\bf d})$, where ${\bf z}$ is the continuous latent variable and ${\bf d}$ is the latent variable associated with the discrete information, labels for example. Then we derive its corresponding ELBO, considering two distinct encoders, characterized by the parameters $\varepsilon$ and $\varsigma$ for the discrete ${\bf z}$ and continuous ${\bf d}$ latent variables, respectively, as follows:
\begin{equation}
\begin{aligned}
&{\mathcal L}^S_{\theta ,\varepsilon ,\varsigma}({\bf x}) = \mathbb{E}_{q_{\varepsilon ,\varsigma } ( {\bf z,d|x})}\log \left[ {\frac{p_\theta ( {\bf x,z,d})}{q_{\varepsilon ,\varsigma } ( {\bf z,d|x})}} \right] \\&= \mathbb{E}_{q_{\varepsilon ,\varsigma } ( {\bf z,d|x})} \log \left[ \frac{p_\theta ( {\bf x|z,d})p ({\bf z}) 
p ( {\bf d} )}{{q_\varepsilon ( {\bf z|x} )q_\varsigma ( {\bf d|x} )}} \right]  \\&
=  \mathbb{E}_{q_{\varepsilon ,\varsigma } ( {\bf z,d|x})} \log [ p_\theta ( {\bf x|z,d} )]  + \mathbb{E}_{q_{\varsigma ,\varepsilon ,\delta } ( {\bf z,d|x})} \log \left[ \frac{p ( {\bf z})}{q_\varepsilon ( {\bf z|x})} \right] \\ 
& + \mathbb{E}_{q_{\varepsilon ,\varsigma } ({\bf z,d|x})}\log \left[ \frac{p ({\bf d})}{q_\varsigma ({\bf d|x})} \right]
= \mathbb{E}_{q_{\varepsilon ,\varsigma } ( {\bf z,d|x})} \log [ p_\theta ( {\bf x|z,d} )] \\&- D_{KL} [ q_\varepsilon ( {\bf z|x} )||p ({\bf z})] - D_{KL}[ q_\varsigma ( {\bf d|x})||p ( {\bf d}) ].
\label{supervised_Elbo}
\end{aligned}
\end{equation}

We assume that ${\bf z}$ is independent from ${\bf d}$, which is guaranteed by using two separate inference models $q_\varepsilon ({\bf z}|{\bf x})$ and  $q_\zeta ({\bf d}|{\bf x})$ for modelling ${\bf z}$ and ${\bf d}$. Eq.~\eqref{supervised_Elbo} corresponds to the ELBO for one component of the mixture model. We then define the mixture's objective function by evaluating a sum over all individual components ELBO's, each multiplied by its associated mixing coefﬁcient~:
\begin{equation}
\begin{aligned}
{\mathcal L}_{L-MVAE}^S({\bf x})  = \sum\limits_{i = 1}^K w_i( \mathbb{E}_{q_{\varepsilon_i,\varsigma_i} ( {\bf z},{\bf d}|{\bf x})}\left[ {\log {p_\theta }\left( {\bf x}|{\bf z},{\bf d} \right)} \right]  \\  -D_{KL} [q_{\varepsilon_i} ({\bf z}|{\bf x})||p ({\bf z}) ] -D_{KL} [ q_{\varsigma_i} ( {\bf d}|{\bf x})||p ({\bf d})]),
\end{aligned}
\label{SuperLoss}
\end{equation}
where $\varepsilon=\{ \varepsilon_i | i=1,\ldots,K  \}$ and $\varsigma =\{ \varsigma_i | i=1,\ldots,K  \}$,  represent the parameters for the encoders modelling continuous ${\bf z}$, and discrete ${\bf d}$, latent variables, for each mixture' component. We call each ${q_{{\xi_i}}}({\bf{d}}|{\bf{x}})$ as the class-specific encoder. The last two terms from (\ref{SuperLoss}) represent the KL divergences between the posterior and prior distributions for the variables ${\bf z}$ and ${\bf d}$, associated to continuous and discrete latent spaces, respectively. 

For the discrete variables we consider sampling using the Gumbel-Max trick for $q_\varsigma ( {\bf d}|{\bf x})$,  as in \cite{Gumble_softmax,GumbelMaxTrick}, in order to produce differentiable discrete variables. We implement $q_\varsigma ( {\bf d}|{\bf x})$ by using a neural network of parameters $\varsigma$ in which the last layer implements the softmax function producing the probability vector $ {\bf d}' = ( d'_1,d'_2,\ldots,d'_K)$, while the sampling process is defined by~:
\begin{equation}
\begin{aligned}
d_k = \frac{ \exp ( ( \log d'_k + g_k)/T)}{\sum\nolimits_{i=1}^K \exp ( (\log d'_i + g_i )/T) } \; ,
\end{aligned}
\end{equation}
where $g_k$ is sampled from the
$\mathop{\rm Gumbel} ( 0,1 )$ distribution. The sample vector ${\bf d}$ is treated as a continuous approximation of the categorical representation (one-hot vector). The sampling process is incorporated into both generation and inference stages. For enforcing the discrete latent variables ${\bf d}$ to capture discriminative information such as the data type, we introduce a mixture of cross-entropy loss ${\cal L}_{S-Mix,\tilde, \varsigma} ({\bf x})$~:
\begin{equation}
\begin{aligned}
{\cal L}_{S-Mix,\varsigma}({\bf x}) = E_{({\bf x},{\bf y})\sim({\cal X},{\cal Y})} \sum\limits_{i = 1}^K w _i \eta(q_{\varsigma_i}({\bf d}|{\bf x}),{\bf y})\,,
\label{CrossEntropy}
\end{aligned}
\end{equation}
where we incorporate the individual VAE components cross-entropy loss $\eta( \cdot,  \cdot ) $ weighted by the associated mixing coefﬁcients, characterizing the encoders specific to learning the discrete variables, into a single objective function for the mixture system and $\varsigma = \{ \varsigma_1,\ldots,\varsigma_k \}$.  The pseudocode for the supervised learning is provided in Algorithm~\ref{A1} where we firstly optimize the parameters of the model by using Eq.~(\ref{SuperLoss}) and Eq.~(\ref{CrossEntropy}) at each iteration.

\noindent \textbf{Lifelong semi-supervised learning.}
We also consider the semi-supervised learning context \cite{Semi_VAE} for the proposed L-MVAE model.  
Under the semi-supervised setting, we only have a small subset of labeled observations $\{{\bf x}_i,y_i|i=1,\ldots,N_{SemS} \}$, with labels $y_i$, with the number of samples $N_{SemS} < N$  and a much larger number of unlabeled data samples 
$\{ \hat{\bf x}_i| i= N_{SemS}+1,\ldots,N \}$ for each learning task, where we assume $N$ data in total, where $\hat{\bf x},{\bf x} \in {\cal X}$. In semi-supervised learning we aim to associate the unlabelled data samples based on their statistical consistency with the labelled data, following model training. Assigned labels $\hat{\bf y}_i$ would then replace discrete variables ${\bf d}$, used for supervised learning, during the decoding process. The objective function for semi-supervised training is~:
\begin{equation}
\begin{aligned}
{\mathcal L}^{SemS}_{Mix,\tilde{\theta},\tilde{\varepsilon} ,\tilde{\varsigma}} (\hat{\bf x}) =  \sum\limits_{i = 1}^K w_i & \left( \mathbb{E}_{q_{\varepsilon_i,\varsigma_i } ( {\bf z},{\bf y}|{\bf x}')}\left[ \log p_\theta \left( {\bf x}'|{\bf z},{\bf y} \right) \right] \right. \\ 
& \left. -D_{KL} [q_{\varepsilon_i} ({\bf z}|\hat{\bf x})||p ({\bf z}) ] \right)
\end{aligned}
\label{SemiSuperLoss}
\end{equation}
where ${\mathcal L}^{SemS}_{Mix,\tilde{\theta},\tilde{\varepsilon} ,\tilde{\varsigma}}(\hat{\bf x})$ is the loss function for the semi-supervised learning of the L-MVAE model, $\sum_{i = 1}^K w_i=1$, while  $\tilde{\theta}$, $\tilde{\varepsilon}$ and $\tilde{\varsigma}$
represent the mixture's model parameters characterizing the decoders and the encoders specific to the continuous ${\bf z}$ and
to the labels ${\bf y}$, respectively.

In addition to ${\mathcal L}_{Mix}^{SemS}(\hat{\bf x})$ from (\ref{SemiSuperLoss}), we also optimize the parameters 
$\tilde{\varsigma}$ using the mixture cross-entropy ${\cal L}_{S-Mix}({\bf x})$, similar to (\ref{CrossEntropy}), used for supervised learning. For the unlabeled samples, missing labels are inferred by using  Gumble-softmax based sampling in which the probability vector ${\bf d}'$ is sampled from the encoder, defined by 
$q_{\varsigma} ({\bf d}|{\bf x}')$. These  discrete variables are then used during the decoding. The final objective function for semi-supervised learning tasks is defined as:
\begin{equation}
\begin{aligned}
{\mathcal L}_{L-MVAE}^{SemS} ({\bf x})  = {\mathcal L}_{Mix}^{SemS} (\hat{\bf x})  + \beta {\mathcal L}_{L-MVAE}^S ({\bf x}) \,,
\label{semi}
\end{aligned}
\end{equation}
where the first term is given in (\ref{SemiSuperLoss}), and $\beta$ controls the importance of the loss associated to the supervised learning ${\mathcal L}_{L-MVAE}^S({\bf x}) $, which is defined in (\ref{SuperLoss}). We separately optimize the parameters of the model by using (\ref{semi}) and (\ref{CrossEntropy}) during each iteration, similar to the supervised learning setting.

\begin{algorithm}
	\caption{Supervised training for the L-MVAE model.}
	\includegraphics[scale=0.69]{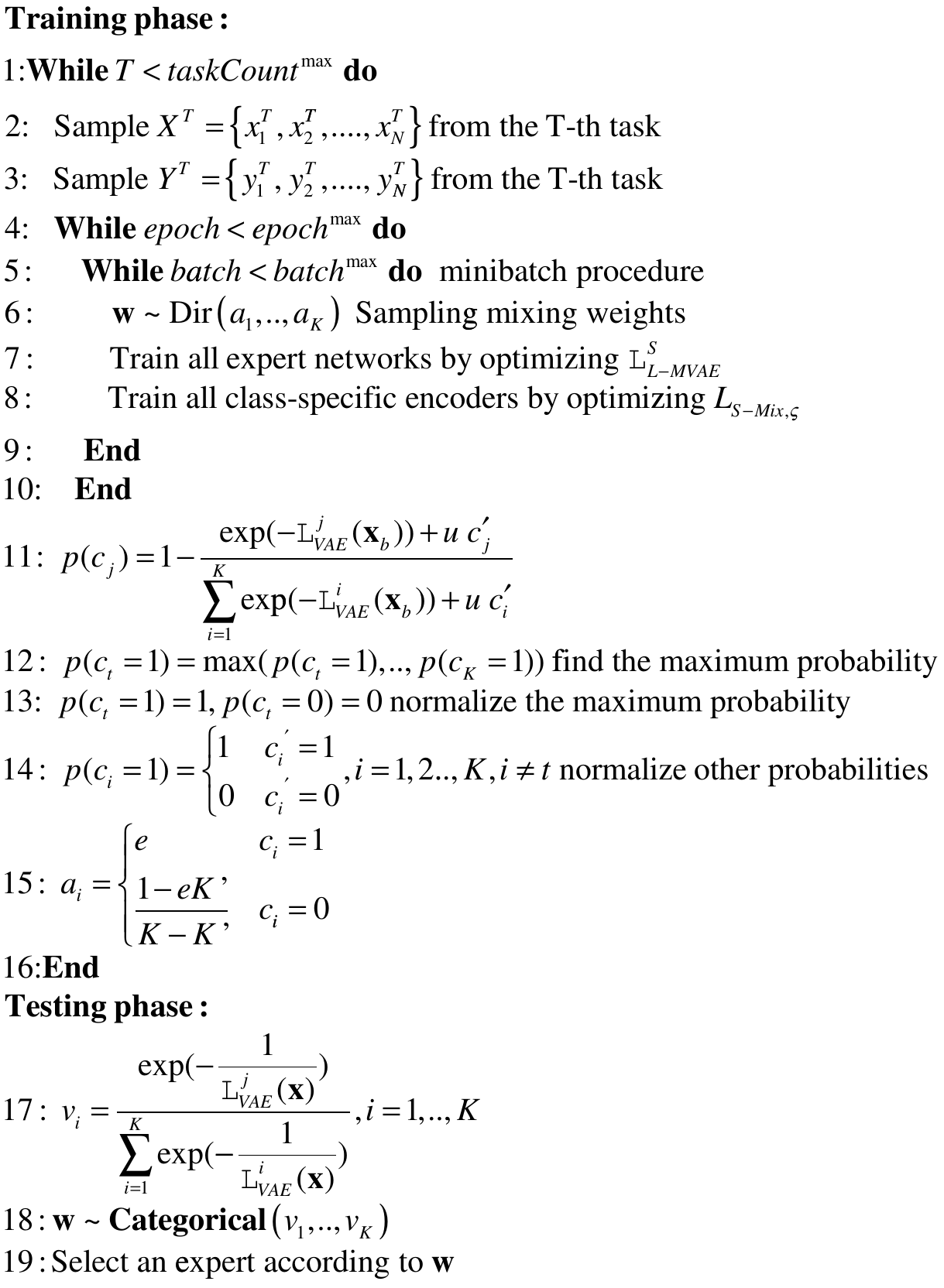}
	\label{A1}
\end{algorithm}

\section{Mixture Expansion Mechanism}
\label{MEM}

\begin{figure*}[htbp]
	\centering
	\includegraphics[scale=0.50]{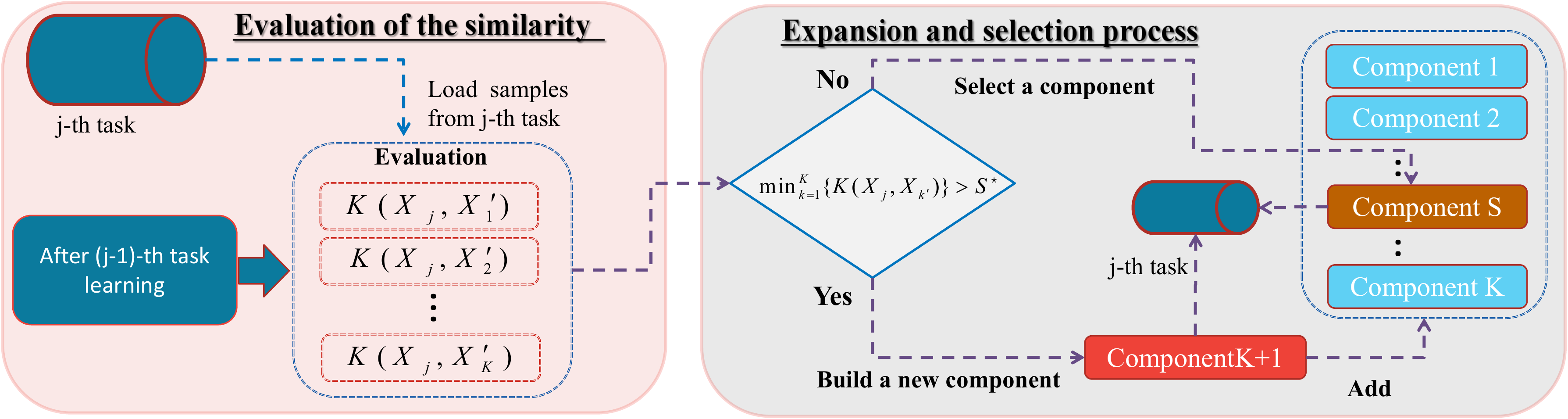}
	\caption{The illustration of the component expansion mechanism. Once learning the $(j-1)$-th task is finished, we collect samples from the $j$-th database and then evaluate the compatibility between this database and the probabilistic representation of each component by using Eq.~\eqref{Criterion}. Then Eq.~\eqref{AddNewComponent} is used for either selecting one of the existing mixture components or expanding the network with a new component (mark withing a red rectangle). The testing phase for the expansion mechanism is identical to the one shown in Fig.~\ref{Mixture_structure}.   }
	\label{Mixture_structure_expansion}
\end{figure*}

A given mixture architecture has limits in its modelling capabilities. Such limits are especially exposed during the lifelong learning, when the model has to learn new tasks. In this section, we introduce a procedure for expanding the L-MVAE architecture in order to enhance the architecture ability to deal successfully with a growing number of tasks. Meanwhile we aim to use a minimal number of model parameters and optimize the training time for efficiently learning all tasks.
We introduce a joint network by adding to the existing VAE component structure consisting of an encoder and a decoder, defined by the parameters ${\theta'_1}$ and  ${\varepsilon'_1}$, respectively, a sub-encoder and a sub-decoder, with parameters $\theta_S$ and $\varepsilon_S$, respectively. During the first task learning, we build the first mixture component based on this joint network. We use $p_{\theta_1}({\bf x}|{\bf z})$ and $q_{\varepsilon_1}({\bf z}|{\bf x})$ to represent the decoder and encoder, respectively, where $\theta_1 = \{ \theta_S,{\theta'_1}\}$ and $\varepsilon_1 = \{ \varepsilon_S,{\varepsilon'_1}\}$. 
During the training we update both the shared parameter set $\{ {\theta _S},{\varepsilon _S}\} $ and the specific parameter set 
$\{ {\theta'_1},{\varepsilon'_1} \} $ when learning the first task. When learning the next task, $\{ \theta_S, \varepsilon_S \} $ parameters are fixed, while a new VAE component is added and only its corresponding specific parameter set  $\{ \theta'_2, \varepsilon'_2 \} $ is updated using Eq.~({\ref{DefineL_VAE}}) using data from the given database. We introduce a new mechanism for acquiring the knowledge corresponding to a new task during the lifelong learning, by either updating an existing mixture component, or adding a new component and training its parameters. The process of the proposed expansion mechanism is shown in Fig.~\ref{Mixture_structure_expansion}.

In order to allow a single component to learn several similar tasks, we introduce a similarity measure between the probabilistic representation associated with a new task and the information recorded by each trained mixture component. If the new task is novel enough relative to the already learnt knowledge, the mixture model will add a new component in order to learn the new task. Otherwise, the training algorithm will select and update the most appropriate component. Let us consider that the mixture model has $K$ components after learning the $j-1$-th task. We 
evaluate the novelty of the $j$-th task by comparing the knowledge acquired by each of the $K$ components and the probabilistic representation of the $j$-th task. We consider a probabilistic representation of the $j$-th task by randomly selecting a set $\{ {\cal X}_j | {\bf x}_{j,i} \in {\cal X}_j; i=1,\ldots,N_j \}$, where in the experiments we consider $N_j=1000$ samples. The probabilistic representation of the knowledge acquired by each expert is represented by its ability to generate specific data. Thus, for each expert $k=1,\ldots,K$, we generate a dataset $\{ {\cal X}'_k | f_{\theta_k} ( {\bf x}_{j,l}) \in {\cal X}_k' ; l=1,\ldots,N_k' \}$, where in the experiments we consider $N_l'=1000$, for $l= 1,\ldots,K$ and $j$ represents the database used for sampling the original data ${\bf x}_{j,l}$. We consider the L2 distance between all data of the two databases, as statistical similarity measure:
\begin{equation}
    {\cal K} ({\cal X}_j , {\cal X}'_k )  = \frac{1}{N_j} \frac{1}{N_k'} \sum_{i=1}^{N_j} \sum_{l=1}^{N_k'} 
    \| {\bf x}_{j,i} - f_{\theta_k} ( {\bf x}_{j,l})  \|\,,
    \label{Criterion}
\end{equation}
for $k= 1,\ldots,K$. A new expert $K+1$ is added to the mixture model when none of the experts is able to generate data similar to those from the new dataset, according to~:
\begin{equation}
 \min_{k=1}^K \{ {\cal K} ({\cal X}_j , {\cal X}_k')\} > S^\star \,,
 \label{AddNewComponent}
\end{equation}
where $S^\star$ is a threshold defining the level of novelty in the knowledge acquired by each expert. The parameter set of the new expert is $\{\varepsilon_S,\varepsilon'_{K+1},\theta_S,\theta'_{K+1}\}$, where only the parameters $\varepsilon'_{K+1},\theta'_{K+1}$ are trained according to the objective defined in ({\ref{DefineL_VAE}}). If (\ref{AddNewComponent}) is not fulfilled then the most suitable component is chosen~:
\begin{equation}
  L =  \arg \min_{k=1}^K {\cal K} ({\cal X}_j , {\cal X}_k') 
 \label{UpdateComponent}
\end{equation}
and its encoder and decoder parameters $\{ {\theta'_L}, {\varepsilon'_L} \} $ are updated. We call the proposed expansion mechanism with the mixture model as L-MVAE dynamic (L-MVAE-Dyn). By considering a fixed component of the model, made up of the sub-decoder and sub-encoder of parameters $\{\varepsilon_S,\theta_S\}$ we ensure a common heritage knowledge for all tasks, corresponding to a set of features shared by the data from several databases. When learning each task, we add an additional set of parameters corresponding to characteristic information from each database. This procedure ensures a fast and efficient learning procedure, while maintaining the required set of parameters to a minimum, when learning a sequence of tasks.

\section{Experiments}
\label{Exper}

We evaluate the performance of the proposed L-MVAE system when learning several tasks. We also assess how L-MVAE is used for semi-supervised and unsupervised learning tasks in the context of lifelong learning. All implementations are done using the TensorFlow framework. 

\subsection{Supervised learning} 

\begin{figure}
    \centering
    \includegraphics[scale=0.50]{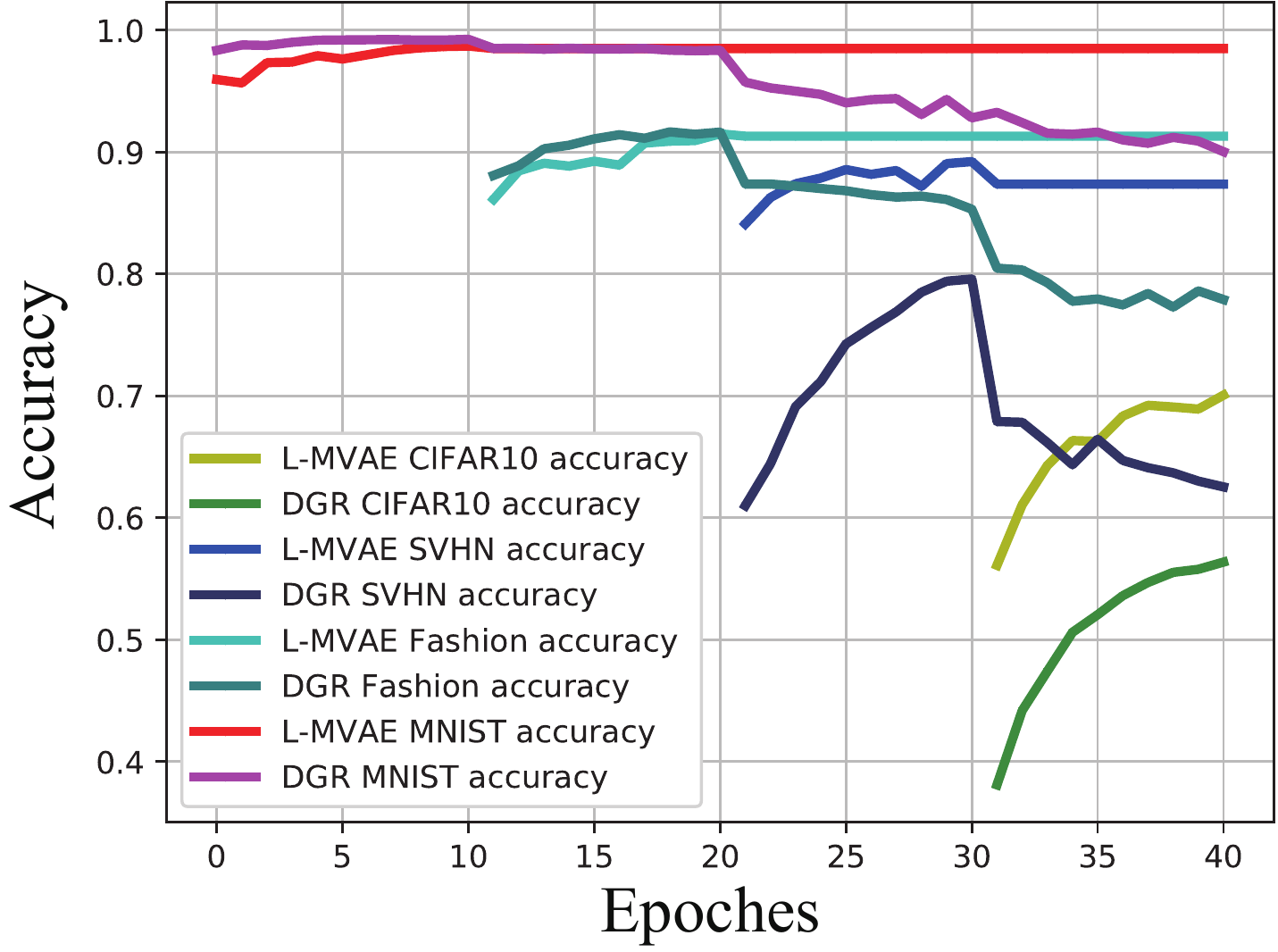}
	\caption{Classification accuracy on all testing data samples across several domains during the lifelong learning.}
	\label{CAccuracy}
\end{figure}

We select four datasets for the lifelong supervised training of L-MVAE: MNIST \cite{MNIST}, Fashion \cite{FashionMNIST}, SVHN \cite{SVHN} and CIFAR10 \cite{CIFAR10}, called MFSC sequence. We estimate the average classification accuracy on all testing data across different domains during the lifelong training, and the results are provided in Fig.~\ref{CAccuracy}. Each task was trained for 10 epochs using Stochastic Gradient Descent (SGD). From these results we observe that each time when training with a new dataset, L-MVAE maintains almost its full performance on the previously learned tasks. For comparison in the same plot from Fig.~\ref{CAccuracy} we show the results obtained by the Deep Generative
Replay (DGR) \cite{Generative_replay} which has a significant performance drop on the previously learnt tasks, when training with a new dataset.

In Table~\ref{accuracyTab} we provide the classification accuracy for the lifelong learning of the MFSC sequence of databases. When all these databases are used jointly for training, within an approach named ``JVAE'', we achieve good results on simple datasets such as MNIST and Fashion, but the performance drops on the datasets containing more complex images. ``Transfer'' represents training a single classifier on a sequence of tasks without using the generative replay mechanism. We can observe that the ``Transfer'' approach only achieves good results on the latest task and completely forgets any previously learnt knowledge. L-MVAE-S is the mixture model sharing the parameters of the decoder with all experts. Although  L-MVAE-S  uses fewer parameters than L-MVAE, it still provides very good results. The generative replay based methods used for comparison, Lifelong  generative modeling (LGM) \cite{GenerativeLifelong}, DGR \cite{Generative_replay} and Continual Unsupervised Representation Learning (CURL) \cite{LifelongUnsupervisedVAE} display a performance drop on all tasks, which is mainly the result of the generative replay samples quality. The generative replay methods tend to forget the previous learnt tasks when learning a sequence of different domains.

\begin{table}[]
    \centering
    \begin{tabular}{lllll}
	\toprule
	\cmidrule(r){1-5}
    Methods  & MNIST     & Fashion &SVHN &Cifar10      \\
	\midrule
   L-MVAE &	\textBF{97.97}&  90.02&  \textBF{87.00}&  \textBF{69.32} \\
     L-MVAE-S &	96.18	&  \textBF{91.64}& 86.20 &  66.94 \\
	JVAEs & 97.72& 88.47 & 61.87 & 52.69  \\
	Transfer & 5.28&  5.23&  13.82& 68.67  \\
	DGR \cite{Generative_replay}  & 90.20&  72.64&  62.44&  56.43 \\
	LGM \cite{GenerativeLifelong} & 61.06&  63.57&  64.21&  56.84  \\
	CURL \cite{LifelongUnsupervisedVAE} & 91.46& 74.29& 66.78& 59.46  \\
	\bottomrule
\end{tabular}
\vspace*{0.2cm}
	\caption{Classification accuracy when considering the lifelong learning of MNIST, Fashion, SVHN and CIFAR10 databases. MFSC and CSFM denote the order of the databases used for the lifelong learning.}
	\label{accuracyTab}
\end{table}

\begin{table*}[h]
	\centering
	\begin{tabular}{lcccccc}
		\toprule
		\cmidrule(r){1-7}
		Dataset    & L-MVAE* &CURL* \cite{LifelongUnsupervisedVAE} & CAE \cite{SemiSuper} & M1 \cite{SemiSuper} &	M1+M2 \cite{SemiSuper} & Semi-VAE \cite{Semi_VAE} \\
		\midrule
	MNIST    &4.95&14.67& 4.77 &4.24 &2.40&2.88   \\
		Fashion    &\textBF{16.93}&64.28& / & /  & / & /    \\
		SVHN    &\textBF{23.00}&66.39& / & / & / & /   \\
		CIFAR10    &\textBF{48.32}&43.57& / & / & / & /   \\
		\bottomrule
	\end{tabular}
	\vspace*{0.2cm}
		\caption{Semi-supervised classification errors on MNIST under the lifelong learning for MNIST, Fashion, SVHN, and CIFAR10 databases.}
	\label{semiTab}
\end{table*}

\subsection{Semi-supervised learning}

We investigate the performance of the L-MVAE in semi-supervised tasks. For the labelled set we randomly select 1,000 training images from the MNIST and 10,000 from each of the datasets: Fashion, SVHN and Cifar10. The remaining data samples are considered as unlabelled. We train the L-MVAE system on both the labelled and unlabelled samples under the MNIST, Fashion, SVHN and Cifar10 lifelong learning, according to Eq.~(\ref{semi}) where we set $\beta {\rm{ = 0}}{\rm{.5}}$. The results are provided in Table~\ref{semiTab}, where we use `*' to denote the model learned under the lifelong setting. The proposed model almost achieves better results than CURL \cite{LifelongUnsupervisedVAE} in each task learning and even achieves competitive results when comparing with the current state of the art semi-supervised methods trained only on a single dataset, such as CAE \cite{SemiSuper}, M1 \cite{SemiSuper}, M1+M2 \cite{SemiSuper} and Semi-VAE \cite{Semi_VAE}.

\subsection{Unsupervised lifelong reconstruction and interpolation}

\begin{figure*}[htbp]
	\centering
	\subfigure[MNIST testing samples]{
		\centering
		\includegraphics[scale=0.45]{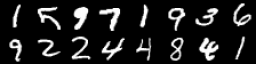}
	}
	\subfigure[Fashion testing samples]{
		\centering
		\includegraphics[scale=0.45]{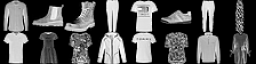}
	}
	\subfigure[SVHN testing samples]{
		\centering
		\includegraphics[scale=0.45]{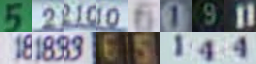}
	}
	\subfigure[CIFAR10 testing samples]{
		\centering
		\includegraphics[scale=0.45]{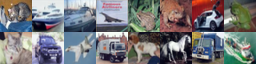}
	}
	\vspace{0.05cm}
	\subfigure[MNIST reconstructions]{
		\centering
		\includegraphics[scale=0.45]{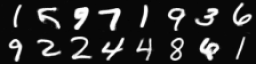}
	}
	\vspace{0.05cm}
	\subfigure[Fashion reconstructions]{
		\centering
		\includegraphics[scale=0.45]{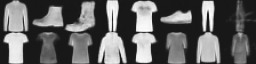}
	}
	\vspace{0.05cm}
	\subfigure[SVHN reconstructions]{
		\centering
		\includegraphics[scale=0.45]{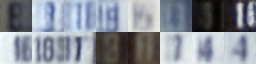}
	}
	\vspace{0.05cm}
	\subfigure[CIFAR10 reconstructions]{
		\centering
		\includegraphics[scale=0.45]{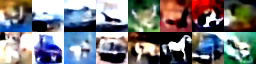}
	}
	\centering
	\caption{Reconstruction results by L-MVAE after the lifelong learning of MNIST, Fashion, SVHN and CIFAR10 (MFSC).}
	\label{Reconstruction32}
\end{figure*}

\begin{figure*}[htbp]
	\centering
	\subfigure[CelebA images.]{
		\centering
		\includegraphics[scale=0.22]{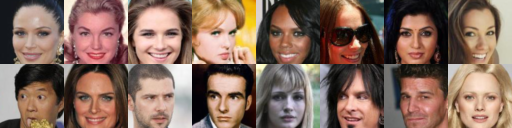}
	}
	\subfigure[CACD images.]{
		\centering
		\includegraphics[scale=0.22]{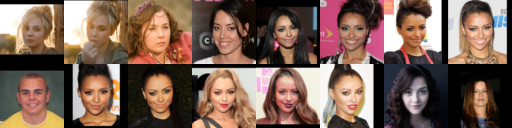}
	}
	\subfigure[3D-chairs images.]{
		\centering
		\includegraphics[scale=0.22]{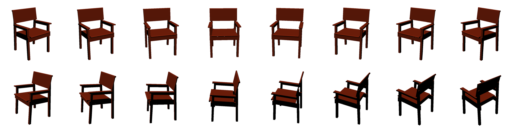}
	}
	\subfigure[Omniglot images.]{
		\centering
		\includegraphics[scale=0.22]{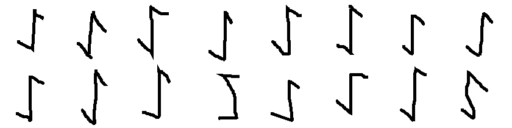}
	}
	\subfigure[CelebA reconstructions.]{
		\centering
		\includegraphics[scale=0.22]{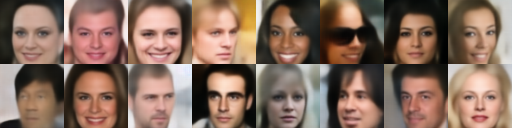}
	}
	\subfigure[CACD reconstructions.]{
		\centering
		\includegraphics[scale=0.22]{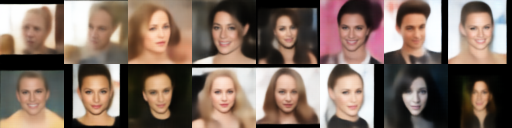}
	}
	\subfigure[3D-chairs reconstructions.]{
		\centering
		\includegraphics[scale=0.22]{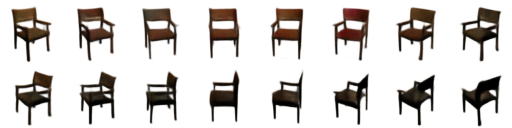}
	}
	\subfigure[Omniglot reconstructions]{
		\centering
		\includegraphics[scale=0.22]{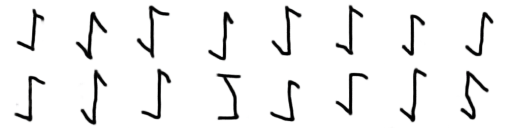}
	}
	\centering
	\caption{ Reconstruction results by L-MVAE after the lifelong learning of the CelebA, CACD, 3D-chairs and Omniglot (CCDO).}
	\label{Reconstruction64}
\end{figure*}

\begin{figure}[htbp]
	\centering
	\subfigure[CelebA interpolation.]{
		\centering
		\includegraphics[scale=0.33]{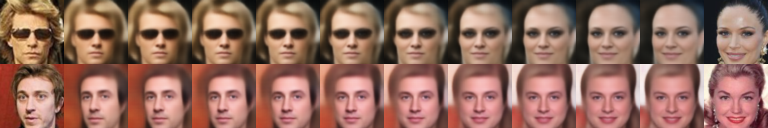}
	}
	\subfigure[CACD interpolation.]{
		\centering
		\includegraphics[scale=0.33]{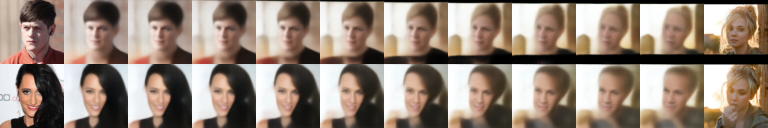}
	}
	\subfigure[3D-chairs interpolation.]{
		\centering
		\includegraphics[scale=0.33]{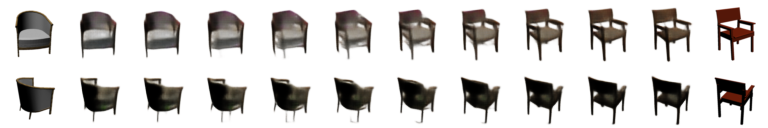}
	}
	\subfigure[Omniglot interpolation.]{
		\centering
		\includegraphics[scale=0.33]{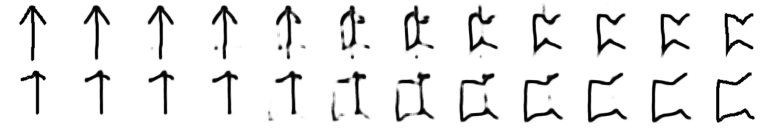}
	} 
	\centering
	\caption{Interpolation results after the lifelong learning of CelebA, CACD, 3D-chairs and Omniglot databases. The images at the ends on each row are real, while those in between are generated by L-MVAE as interpolations exploring the latent space.}
	\label{Interpolation}
\end{figure}

In the following, L-MVAE model is used in unsupervised applications, where there are no data labels. We train the proposed mixture system with four components ($K=4$) under the MNIST, Fashion, SVHN and Cifar10 (MFSC) as well as when using CelebA, CACD, 3D-chairs and Omniglot (CCDO) lifelong learning settings. The original images for MFSC and for CCDO databases are provided in Figures~\ref{Reconstruction32} a-d and \ref{Reconstruction64} a-d, respectively.
The image reconstruction results corresponding to these images, following the lifelong learning, are shown in Figures~\ref{Reconstruction32} e-h, and Figures~\ref{Reconstruction64} e-h, respectively. These results show that the proposed L-MVAE mixture system is able to make accurate inference across several different domains.
We also perform interpolations in the latent space of multiple domains. When interpolating between two latent vectors, we initially select the most relevant expert, according to the selection strategy from Section~\ref{SelectExpert}, and then infer the latent variables using the selected inference model. The selected decoder will then recover images from the interpolated latent variable space. We present the interpolation results in Figures~\ref{Interpolation} a-d, for images from CelebA, CACD, 3D-chairs and Omniglot databases. The proposed model achieves continuity in the latent space as reflected in the generated images derived by each expert, according to these results.

\subsection{Disentangled representation learning}

\begin{figure}
    \centering
    \subfigure[Age change]{
		\centering \hspace*{-1cm}
		\begin{minipage}{8.0cm}
		\includegraphics[scale=0.33]{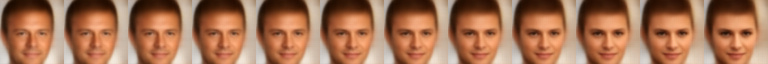}
		\includegraphics[scale=0.33]{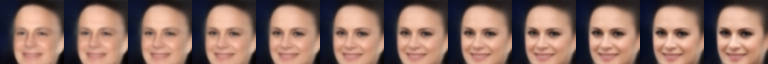} 
	   \end{minipage}
	}
	\subfigure[Hair style]{
		\centering \hspace*{-1cm}
		\begin{minipage}{8.0cm}
			\includegraphics[scale=0.33]{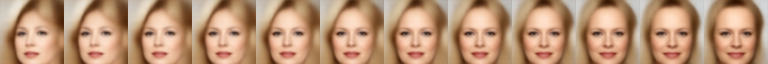}
			\includegraphics[scale=0.33]{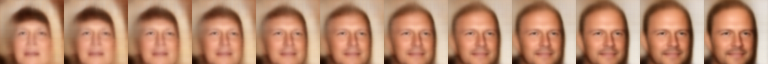} 
		\end{minipage}
	}
	\subfigure[Illumination]{
		\centering \hspace*{-1cm}
		\begin{minipage}{8.0cm}
			\includegraphics[scale=0.33]{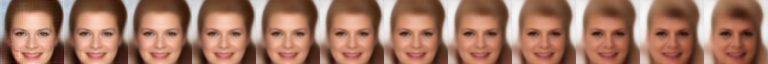}
			\includegraphics[scale=0.33]{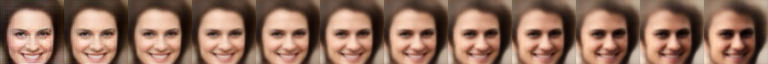} 
		\end{minipage}
	}
	\subfigure[Orientation]{
		\centering \hspace*{-1cm}
		\begin{minipage}{8.0cm}
			\includegraphics[scale=0.33]{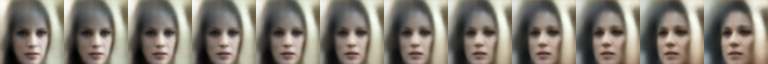}
			\includegraphics[scale=0.33]{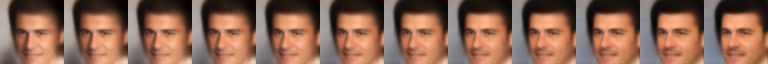} 
		\end{minipage}
	}
	\centering
	\caption{Results, showing disentanglement, by changing specific properties of image representations, after the Lifelong training with CelebA, CACD, 3D-chairs and Omniglot databases using the loss function from (\ref{UnDisentangled_loss}) .}
	\label{Disentangled}
\end{figure}

\begin{figure}
    \centering
	\subfigure[Chair size]{
		\centering 	\hspace*{-1cm}
		\begin{minipage}{8.0cm} 
			\includegraphics[scale=0.33]{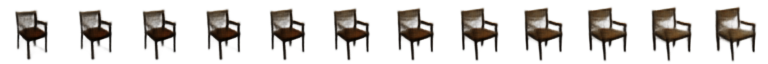}
			\includegraphics[scale=0.33]{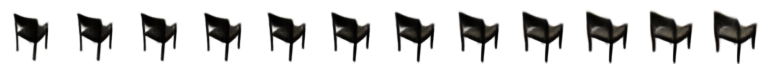} 
		\end{minipage}
	}
	\subfigure[Chair style]{
	\centering 	\hspace*{-1cm}
	\begin{minipage}{8.0cm} 
		\includegraphics[scale=0.33]{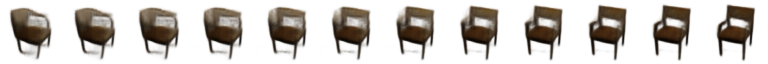}
		\includegraphics[scale=0.33]{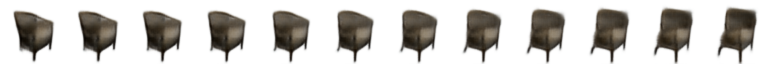} 
	\end{minipage}}
	\subfigure[Chair orientation]{
		\centering \hspace*{-1cm}
		\begin{minipage}{8.0cm} 
			\includegraphics[scale=0.33]{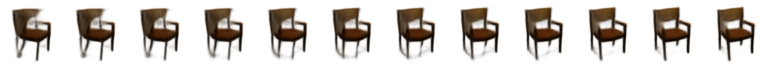}
			\includegraphics[scale=0.33]{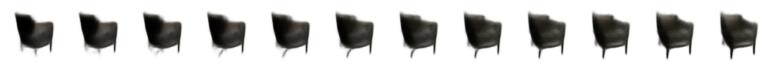} 
	\end{minipage}}
	\centering
	\caption{Results, showing disentanglement, by changing specific properties of image representations, after the Lifelong training with CelebA, CACD, 3D-chairs and Omniglot databases.}
	\label{Disentangled2}
\end{figure}

We train L-MVAE system under the CelebA, CACD, 3D-chairs and Omniglot lifelong learning by using the disentangled loss function from Eq.~(\ref{UnDisentangled_loss}) where $C$ is increased from a very small value to 25.0 during the training and we set $\gamma {\rm{ = 4}}$. After the training, the L-MVAE system firstly chooses the most relevant expert and then a single latent variable, inferred by the selected expert, is changed from -3 to 3 while fixing the other latent variables. The results are shown in Figures~\ref{Disentangled} and \ref{Disentangled2}. From Figures~\ref{Disentangled} a-d we observe that the proposed L-MVAE approach can discover four disentangled representations for CelebA by changing: age, hair style, illumination and face orientation. From Figures~\ref{Disentangled2} a-c we can observe that we can change chair size, style and orientation.

\begin{table}[]
    \centering
    \begin{tabular}{lcccl}
		\toprule
		\cmidrule(r){1-5}
		Dataset & L-MVAE & CURL \cite{LifelongUnsupervisedVAE} & JVAE  & Lifelong  \\
		\midrule
		MNIST & \textBF{48.66} &272.95&195.79  &MFSC  \\
		Fashion & \textBF{53.02}&190.36 &173.64  &MFSC  \\
		SVHN & \textBF{40.39}&127.64 &208.19  &MFSC  \\
		Cifar10 & \textBF{752.91}&1409.74 &1840.40  &MFSC  \\
		MNIST & \textBF{43.26}&64.99 & /  &CSFM  \\
		Fashion & \textBF{44.15}&131.13 & /&CSFM  \\
		SVHN & \textBF{39.46} &278.01 & / &CSFM  \\
		Cifar10 & \textBF{778.17}&2406.13 & / &CSFM  \\
		\bottomrule
	\end{tabular}
	\vspace*{0.2cm}
		\caption{Negative log-likelihood (NLL) estimation for all testing images for the lifelong learning of the probabilistic representations for MNIST, Fashion, SVHN and CIFAR10 databases.}
	\label{rTab}
\end{table}

\begin{table}[]
    \centering
	\begin{tabular}{lcccl}
		\toprule
		\cmidrule(r){1-5}
		Dataset & L-MVAE & CURL \cite{LifelongUnsupervisedVAE} & JVAE  &Lifelong  \\
		\midrule
		MNIST & \textBF{25.83} &167.09&68.59  &MFSC  \\
		Fashion & \textBF{34.09}&139.91 &141.16  &MFSC  \\
		SVHN & \textBF{27.20}&84.45 &295.94  &MFSC  \\
		Cifar10 & \textBF{631.14}&1225.41 &1792.08  &MFSC  \\
		MNIST & \textBF{20.09}&33.55 & /  &CSFM  \\
		Fashion & \textBF{26.46}&252.53 & /&CSFM  \\
		SVHN & \textBF{25.81} &110.21 & / &CSFM  \\
		Cifar10 & \textBF{653.39}&2340.37 & / &CSFM  \\
		\bottomrule
	\end{tabular}
	\vspace*{0.2cm}
		\caption{Image average reconstruction error for after the lifelong learning of MNIST, Fashion, SVHN and CIFAR10 databases.}
	\label{rTab6}
\end{table}

\begin{figure*}[htbp]
\centering
	\subfigure[Reconstruction errors when changing the number of experts.]{
		\centering
	\includegraphics[scale=0.33]{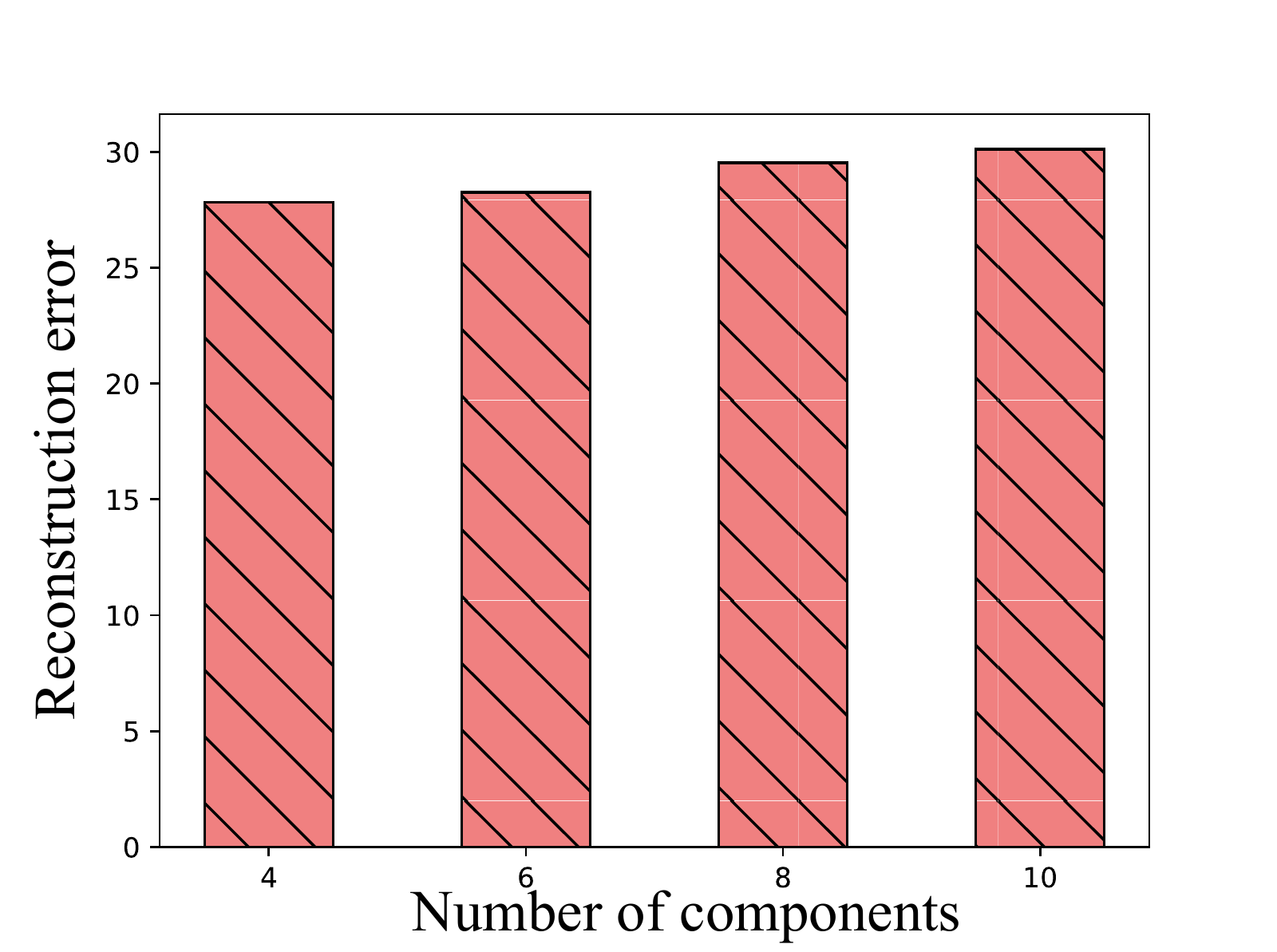}
	}
	\centering
	\subfigure[NLL estimation when selecting mixture components during the training and without component dropout.]{
		\centering
	\includegraphics[scale=0.33]{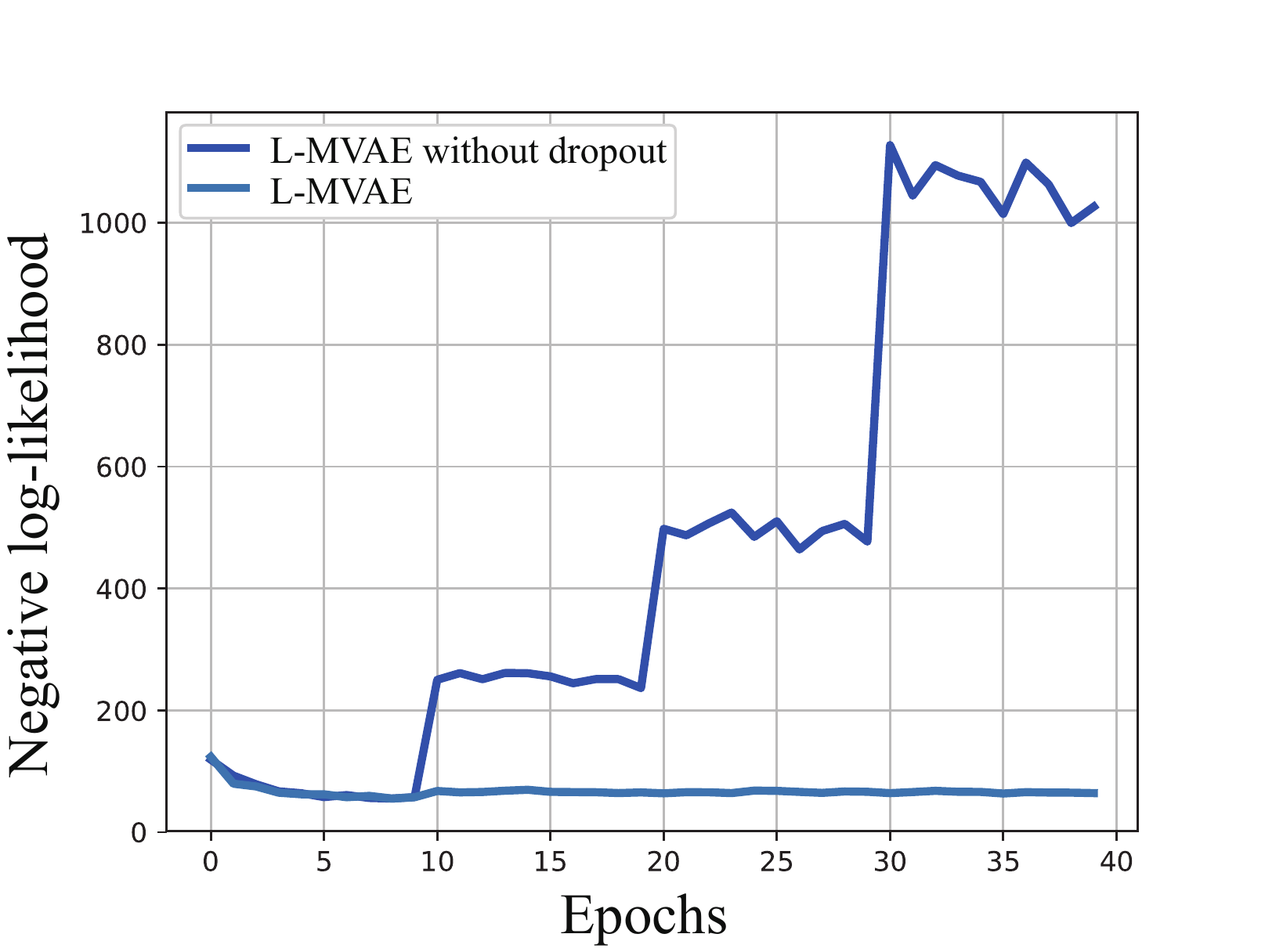}
	}
	\centering
	\subfigure[MELBO and ELBO estimation.]{
		\centering
		\includegraphics[scale=0.33]{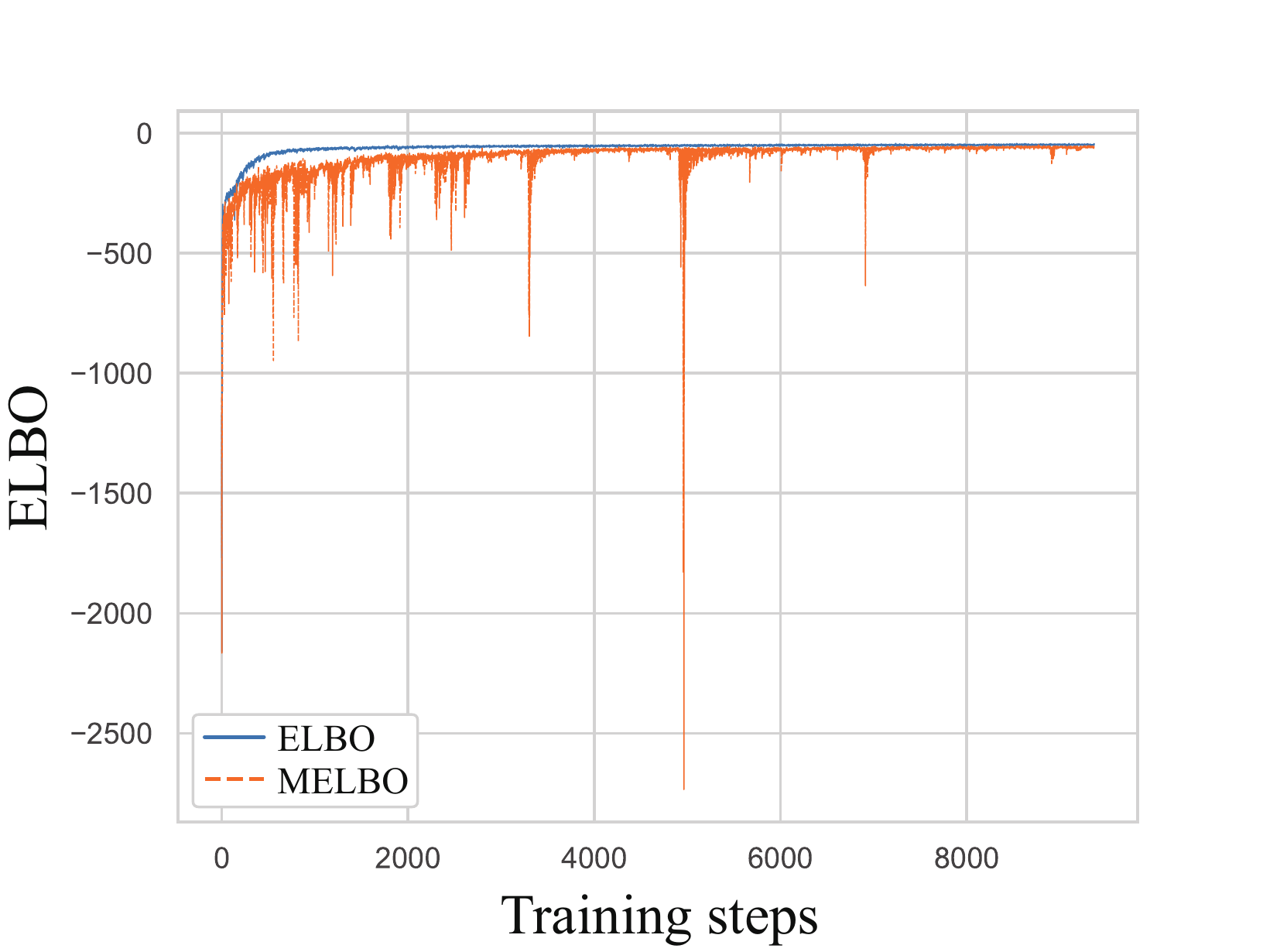}
	}
	\caption{Analysis results for the L-MVAE framework.}
	\label{avgRecoAcrossModels}
\end{figure*}

\begin{figure}[htbp]
	\centering
		\includegraphics[scale=0.5]{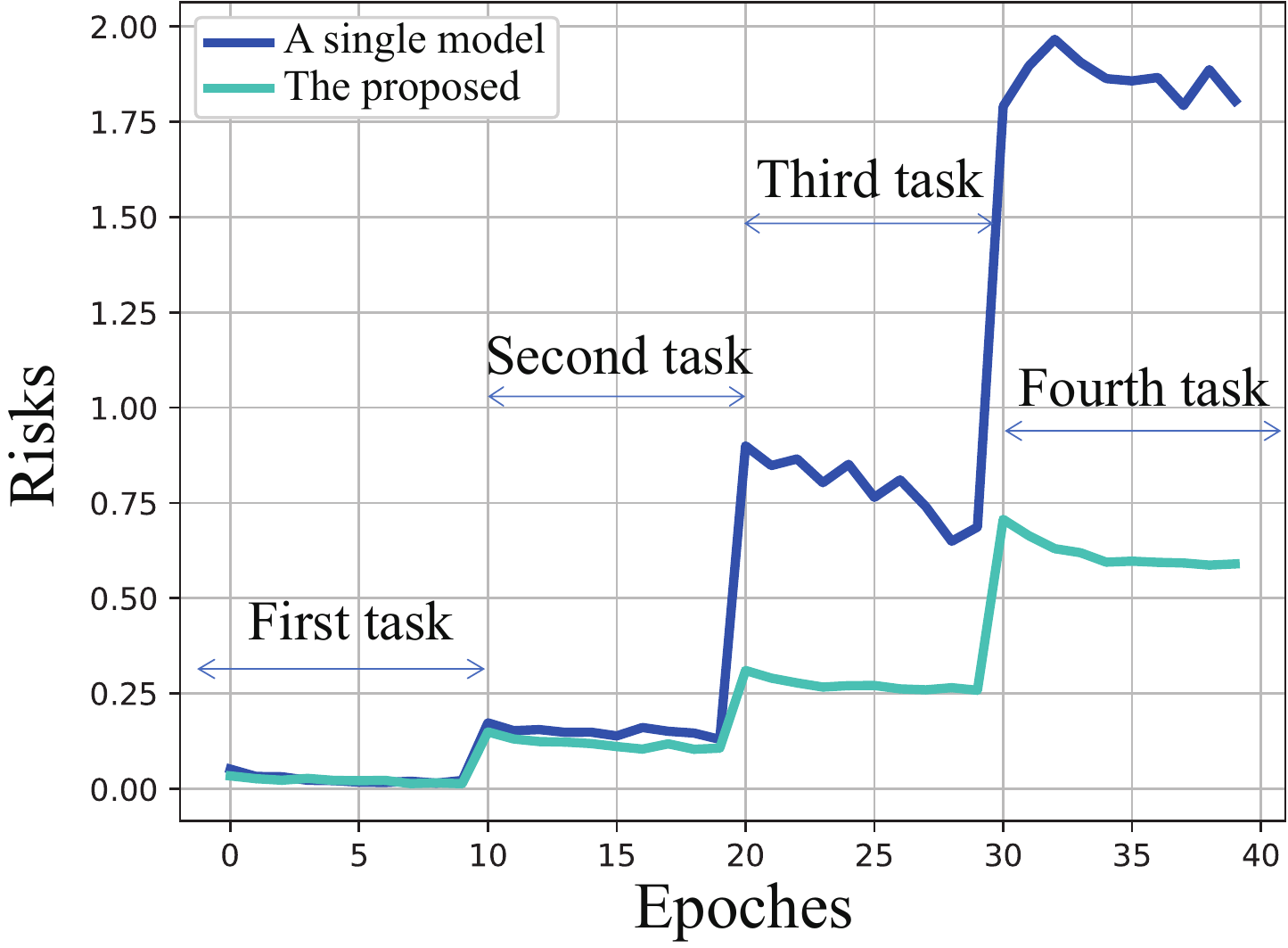}
	\centering
	\caption{ The risks evaluated at each epoch under the MNIST, Fashion, SVHN and CIFAR10 lifelong learning. }
	\label{theory2}
\end{figure}

\begin{figure}[htbp]
	\centering
	\includegraphics[scale=0.5]{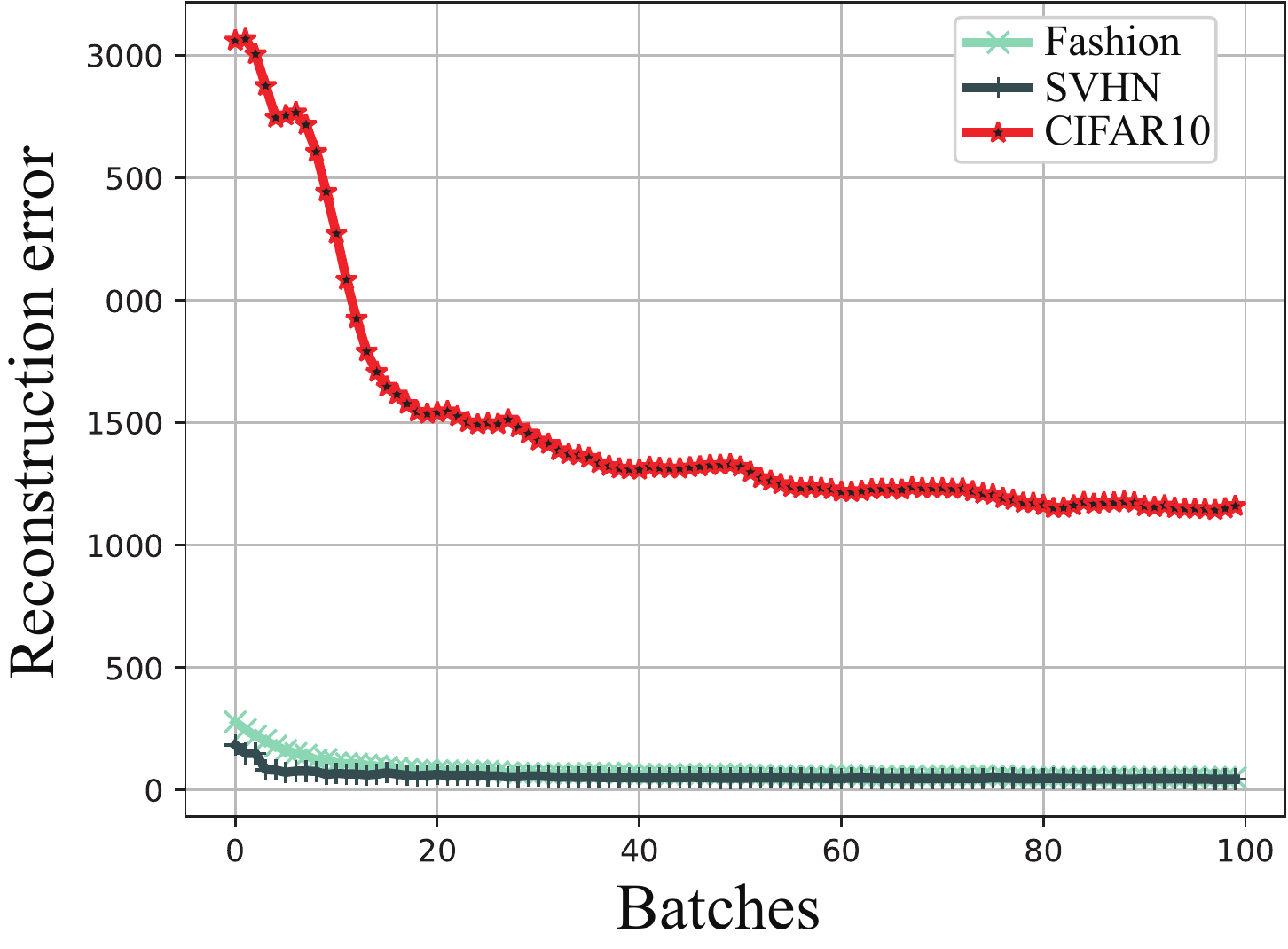}
	\centering
	\caption{The transfer learning ability $s({\theta_k,j})$, defined in Eq.~(\ref{TransfAbility}), for L-MVAE under the MNIST, Fashion, SVHN and CIFAR10 lifelong learning.  }
	\label{transfer}
\end{figure}

\begin{figure*}[htbp]
	\centering
	\subfigure[Fashion]{
		\centering
		\includegraphics[scale=0.34]{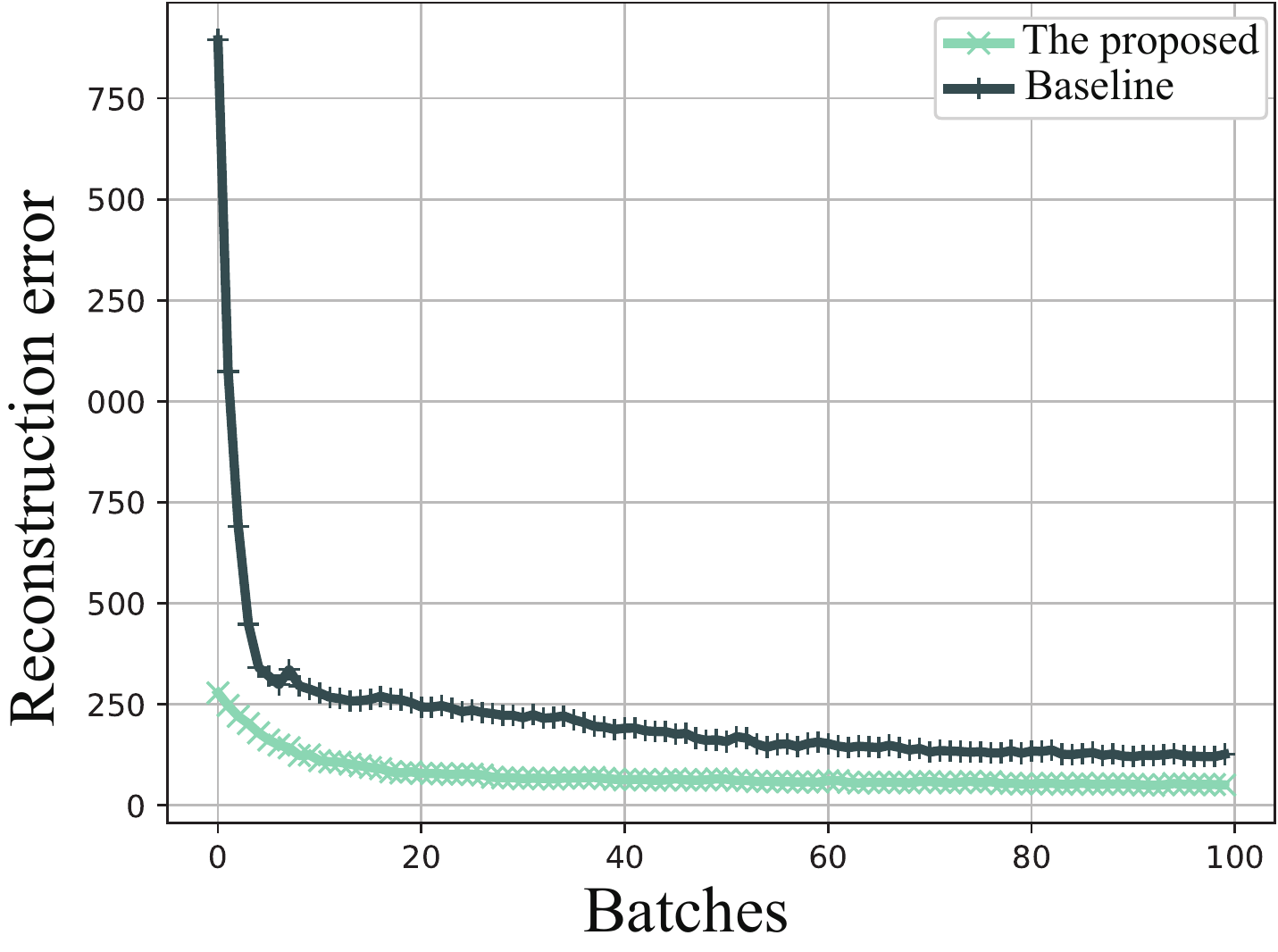}
	}
	\subfigure[SVHN]{
		\centering
		\includegraphics[scale=0.34]{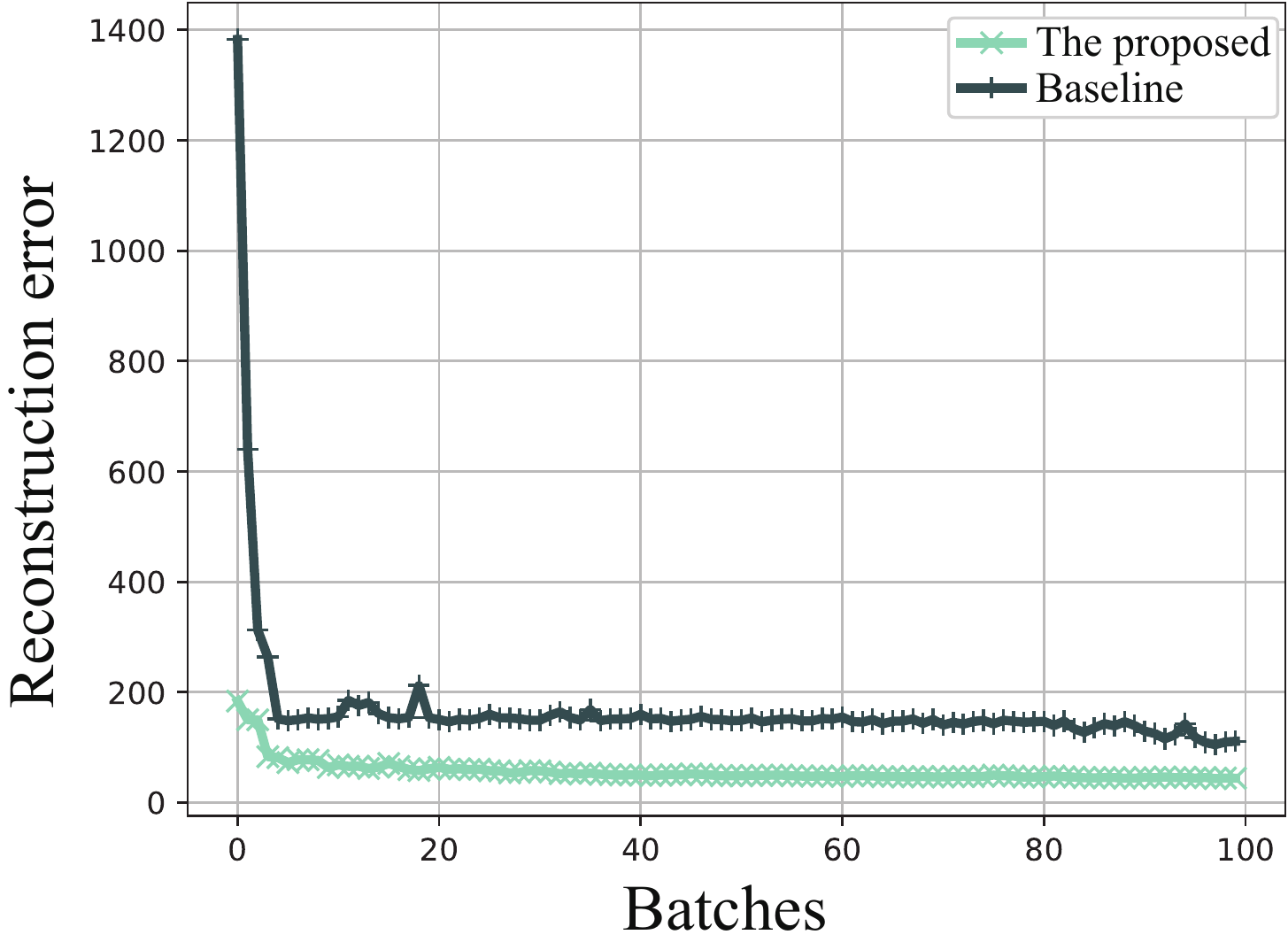}
	}
	\subfigure[Cifar10]{
		\centering
		\includegraphics[scale=0.34]{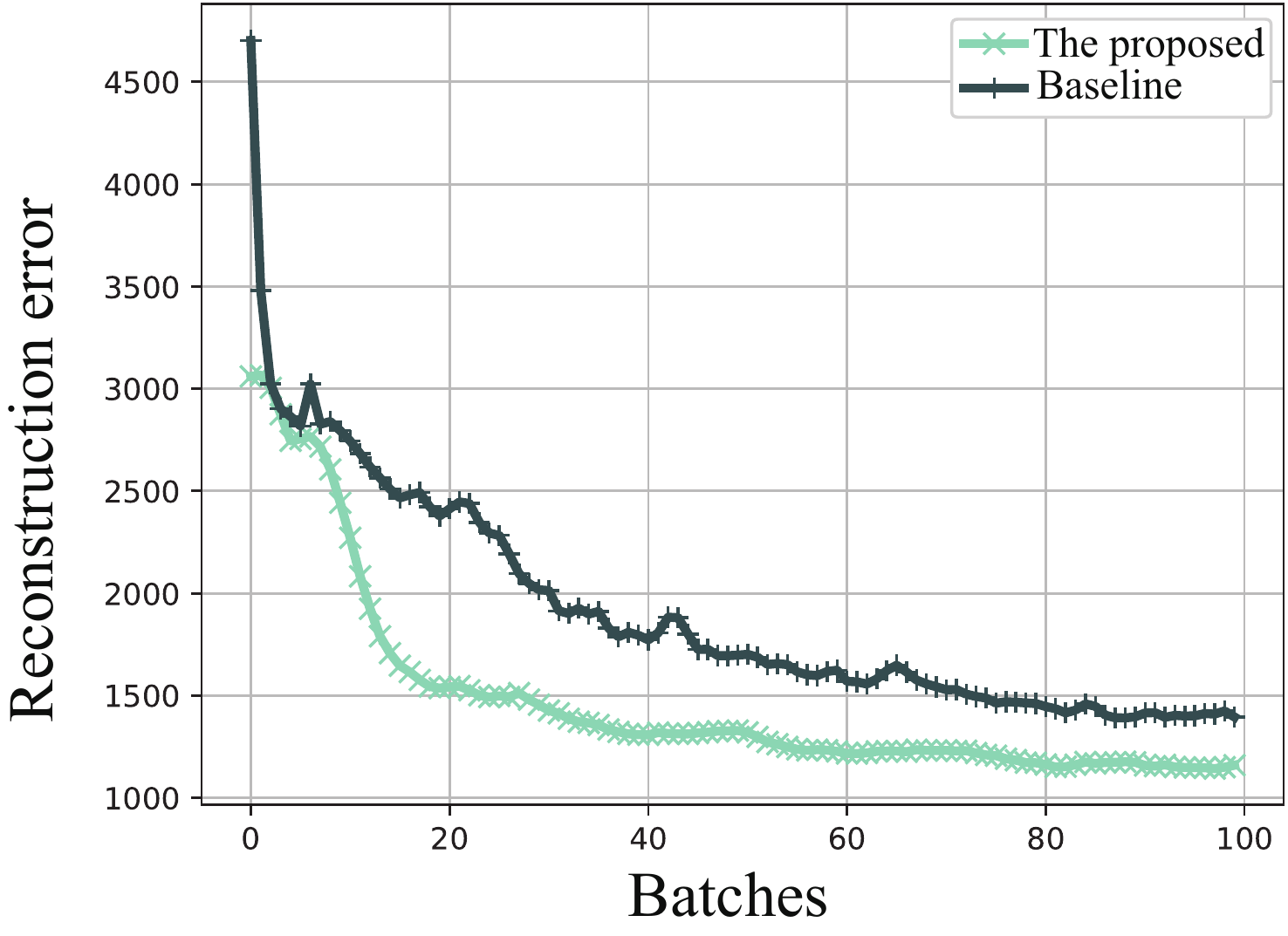}
	}
	\centering
	\caption{ Average reconstructions errors during the lifelong learning. }
	\label{transfer2}
\end{figure*}

\subsection{Visual quality evaluation for the generated images}

For assessing the representation learning ability under the lifelong setting, we evaluate the negative log-likelihood (NLL), representing the reconstruction error plus the KL divergence term, as well as we evaluate the inception score (IS) \cite{InceptScore} for the reconstructed images from the testing set. First, we train various models under the MNIST, Fashion, SVHN and CIFAR10 (MFSC) lifelong learning setting, by considering 100 epochs for learning each task. The results for MFSC and when considering the learning of the databases in reversed order as CSFM, are provided in Tables~\ref{rTab} and \ref{rTab6} for the average NLL and the average reconstruction error, respectively. These results show that the proposed approach achieves the best results when compared with CURL \cite{LifelongUnsupervisedVAE}, LGM \cite{GenerativeLifelong} and with JVAE (when training with all databases at once).

We also consider the lifelong training for  ImageNet, CIFAR100, CIFAR10 and MNIST. After the training, we choose 5,000 images for testing from CIFAR10, CIFAR100 and ImageNet, respectively, while the IS score of the reconstructed images is provided in Table~\ref{ISTab1} when comparing with CURL \cite{LifelongUnsupervisedVAE} and LGM \cite{GenerativeLifelong}. Then we train various models under the CIFAR100, CIFAR10 and ImageNet lifelong learning and we provide the results in Table~\ref{ISTab2}. These results show that the proposed model still provide the best performance even when learning a sequence of several databases containing complex and diverse images. 

\begin{table}[]
    \centering
    \begin{tabular}{lccc}
		\toprule
		\cmidrule(r){1-4}
		Dataset & L-MVAE & CURL \cite{LifelongUnsupervisedVAE} & LGM \cite{GenerativeLifelong}   \\
		\midrule
		CIFAR10 & \textBF{4.82} &3.85&3.23   \\
		CIFAR100 & \textBF{4.78}&3.56 &3.64   \\
		ImageNet & \textBF{5.01}&3.72 &3.47   \\
		\bottomrule
	\end{tabular}
	\vspace*{0.2cm}
		\caption{The IS score for 5,000 testing images under the lifelong learning of ImageNet, CIFAR100, CIFAR10 and MNIST databases.}
	\label{ISTab1}
\end{table}

\begin{table}[]
    \centering
    \begin{tabular}{lcc}
		\toprule
		\cmidrule(r){1-3}
		Dataset & L-MVAE & CURL \cite{LifelongUnsupervisedVAE}    \\
		\midrule
		CIFAR10 & \textBF{4.73} &3.59   \\
		CIFAR100 & \textBF{4.13}&3.47   \\
	ImageNet & \textBF{5.52}&3.56   \\
		\bottomrule
	\end{tabular}
	\vspace*{0.2cm}
		\caption{The IS score for generated images after the lifelong learning of CIFAR100, CIFAR10 and ImageNet databases.}
	\label{ISTab2}
\end{table}

\subsection{Ablation study}

We perform an ablation study to investigate the performance when we change the configuration of the mixture model. We train L-MVAE with $K \in \{4, 6, 8, 10\}$ components under the MNIST, Fashion, SVHN and Fashion lifelong learning setting. We plot the average reconstruction errors on all MNIST testing samples in Fig.~\ref{avgRecoAcrossModels}a. The results show that the number of components does not affect the performance too much and this is why we use $K=4$ components in the experiments. 

We also investigate the performance of the proposed model when not properly estimating the Dirichlet parameters, where the weights $w_i$, $i=1,\ldots,4$ are sampled from the same distribution. We call the model that does not have a component selection as "L-MVAE without dropout". We train this model under the same lifelong task learning as above and the NLL results on the first task (MNIST) are shown in Fig.~\ref{avgRecoAcrossModels}b, where it can be observed that this model would lose its performance during the following tasks when not following the  dropout approach described in Section~\ref{SelectExpert}.  These results are because all experts are activated during the learning of the following tasks if the Dirichlet parameters are not changed accordingly.

In the following experiments we train the L-MVAE model under the lifelong learning of MNIST, Fashion, SVHN and CIFAR10, where we evaluate MELBO, from Eq.~(\ref{MELBO}), for each training step in the first task and the results are shown in Fig.~\ref{avgRecoAcrossModels}c where we also consider a single VAE model with optimal ELBO training on MNIST (MELBO and ELBO are estimated by using the negative reconstruction errors and KL divergence). From these results, MELBO is always bounded by this optimal ELBO and still represents a lower bound on the sample log-likelihood since $\log p({\bf x}) \ge$ ELBO, according to {\em Theorem 2} from Section~\ref{TheoAnal}. We also train a single expert with GRM and a mixture model with 4 experts under MNIST, Fashion, SVHN and CIFAR10 lifelong learning. We consider the classification error rate as the risk of a model evaluated on the testing set and the accumulated errors are calculated by summing up the risks on the testing sets of all learnt tasks. We consider 10 epochs for each task training and plot the results in Fig.~\ref{theory2}. We observe that when considering a single model tends to have a large risk while learning additional tasks. The proposed L-MVAE mixture model always has a lower risk than a single VAE.
 
\begin{figure}[htbp]
	\centering
		\includegraphics[scale=0.45]{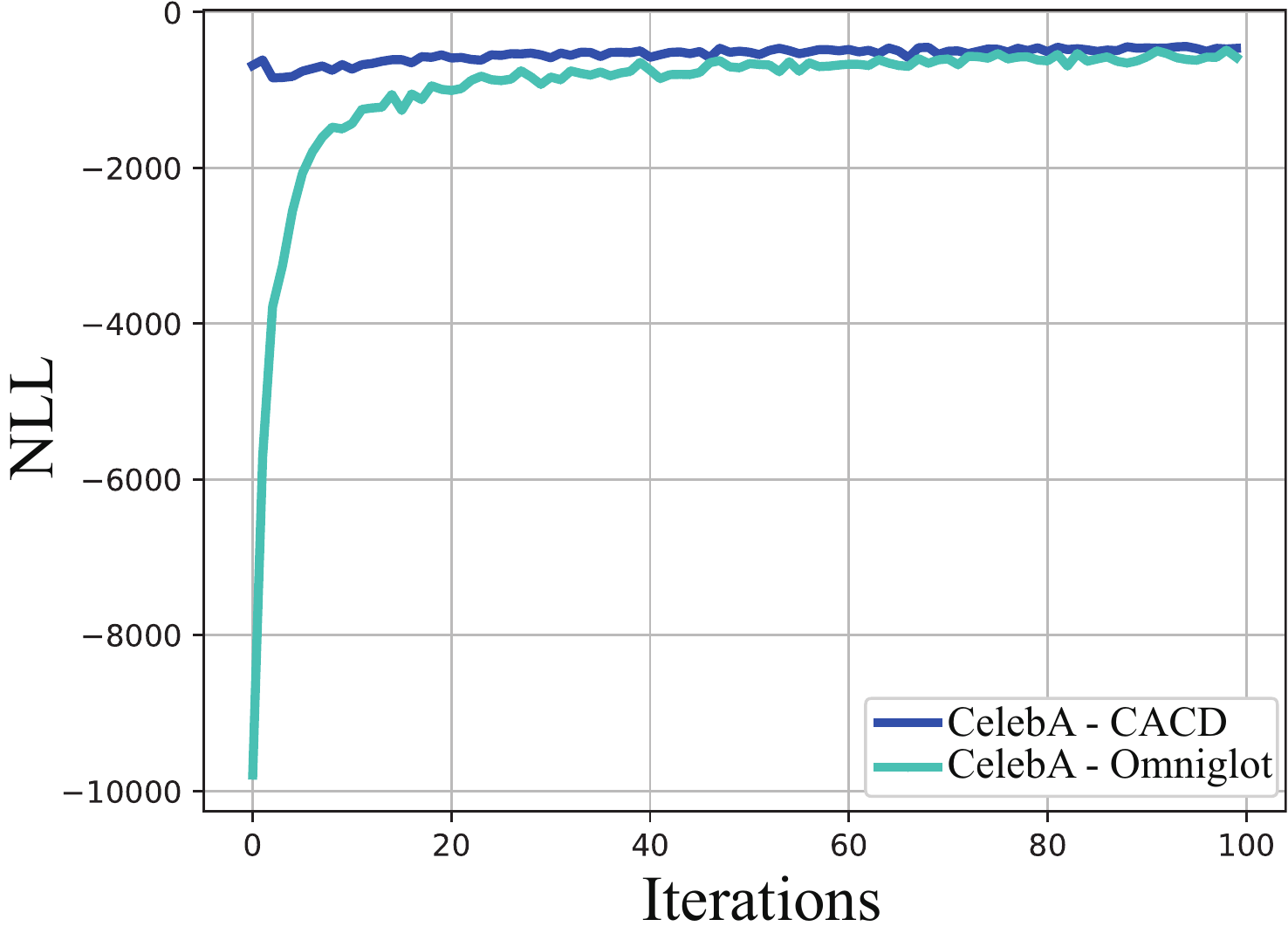}
	\centering
	\caption{ The negative log-likelihood (NLL) evaluated when learning the second task. }
	\label{transfer3}
\end{figure}

\begin{table}[h]
	\centering
	\begin{tabular}{llllc}
		\toprule
		Model & Lifelong &  {\em IS} \\
		\midrule
	MIX+Wasserstein GAN in \cite{MixWass} & No & 4.04 \\
		DCGAN \cite{ConvGAN} in \cite{VAE_symmetric} & No &4.89 \\
		ALI \cite{AdLearnInf} in \cite{VAE_symmetric} & No &4.97 \\
		PixelCNN++ \cite{Pixelcnn++} in \cite{ASymVAE} & No & 5.51 \\
		WGAN in \cite{MixWass} & No &3.82 \\	
		L-MVAE-MFSC & Yes &5.77 \\
		L-MVAE-CSFM & Yes &5.322 \\
		\bottomrule
	\end{tabular}
	\vspace*{0.2cm}
	\caption{Inception Score (IS) evaluated on CIFAR10.}
	\label{Tab5}
\end{table}

\begin{table*}[t]
\centering 
\begin{tabular}{l c c c c cc c} 
\toprule 
& \multicolumn{2}{c}{\textbf{MSE}}&
\multicolumn{2}{c}{\textbf{SSMI}}& \multicolumn{2}{c}{\textbf{PSNR}} \\ 
\cmidrule(l){2-3}\cmidrule(l){4-5}\cmidrule(l){6-7} 
\textbf{Datasets} &  L-MVAE-Dynamic & BatchEnsemble \cite{BatchEnsemble} & L-MVAE-Dynamic& BatchEnsemble \cite{BatchEnsemble} & L-MVAE-Dynamic& BatchEnsemble \cite{BatchEnsemble} 
\\ 
\midrule 
MNIST & 20.45 &19.06&0.91 & 0.92&22.24 & 22.64
		\\
		Fashion & 35.55 &179.72&0.74 &0.27&19.45 &12.23
		\\
		SVHN& 31.78 &130.63&0.63 &0.35&15.37 &9.05
		\\
		CIFAR10& 853.42  &846.52&0.34 &0.36&17.28 &17.34
		\\
\midrule
\midrule
		Average &\textBF{235.30} &293.98&\textBF{0.66} &0.47&\textBF{18.59}  &15.31
		\\
\bottomrule 
\end{tabular}
\vspace{3pt}
		\caption{The reconstruction performance of various models under the MFSC lifelong learning.}
	\label{Unsupervised2}
\end{table*}

\begin{table}[]
    \centering
    \begin{tabular}{lcc}
		\toprule
		\cmidrule(r){1-3}
		Dataset & L-MVAE-Dynamic & BatchEnsemble \cite{BatchEnsemble}  \\
		\midrule
		MNIST & 99.18& 99.34 \\
		Fashion &90.46&88.52 \\
	SVHN &78.63&74.85  \\
		CIFAR10 &63.71&54.80  \\
	\midrule
		\midrule
		Average &\textBF{82.99} &79.38 \\
		\bottomrule
	\end{tabular}
	\vspace*{0.2cm}
		\caption{The classification accuracy of various models under MFSC lifelong learning.}
	\label{acc_table}
\end{table}

\begin{table}[]
    \centering
    \begin{tabular}{lcc}
		\toprule
		\cmidrule(r){1-3}
		Dataset & L-MVAE-Dynamic & BatchEnsemble \cite{BatchEnsemble}  \\
		\midrule
		MNIST&89.40 &99.17 \\
		SVHN&86.54 &70.59 \\
		Fashion&90.44&85.72 \\
		IFashion&89.35&85.47 \\
		IMNIST&99.27&98.84 \\
		RFashion&91.74&87.39 \\
	    CIFAR10&67.91&54.69 \\
	\midrule
		\midrule
		Average&\textBF{87.81}&83.13  \\
		\bottomrule
	\end{tabular}
	\vspace*{0.2cm}
		\caption{The classification accuracy of various models under MSFIIRC lifelong learning.}
	\label{long_acc_table}
\end{table}

\subsection{Transfer metric and transfer learning}

In this section, we evaluate how quickly L-MVAE  learns a new task when presented with a new database for training. The learning of the probabilistic representation of a new dataset by L-MVAE, can be interpreted as a knowledge transfer process from one domain to another. This results in mixing the information being learnt by the expert from the new database with the information already stored in the networks' parameters, corresponding to the previously learnt tasks. In this paper, we propose a new metric, assessing the ability for transferring information during the lifelong learning when learning each new task~:
\begin{equation}
\begin{aligned}
s({\theta_k,j}) = \frac{1}{N_j}\sum\limits_{i = 1}^{N_j} \delta
( {\bf x}_{i,j},f_{\theta_k} ( {\bf x}_{i,j})) 
\end{aligned}
\label{TransfAbility}
\end{equation}
where $s({\theta_k,j})$ is the performance score of the $k$-th mixture component of parameters $\theta_k$ for the $j$-th task, and $\{ {\bf x}_{i,j} \in {\cal X}_j | i=1,\ldots,N_j\}$ represents a given batch of images sampled from the $j$-th database, and  $\delta (\cdot,\cdot) $ is the performance metric, considered as either the Mean Square Error (MSE), or it can be the classification accuracy, depending on the application of each task. $f_{\theta_k}( {\bf x}_{i,j})$ represents the image  reconstructed by the L-MVAE model considering the given batch of images ${\bf x}_{i,j}$ corresponding to the $j$-th task. The proposed metric can measure the training efficiency when a model is trained with a new task, representing the information transfer ability of the model when learning new tasks. 

We train the L-MVAE model under the MNIST, Fashion, SVHN and CIFAR10 lifelong learning setting. The transfer learning ability during the lifelong learning is evaluated in Fig.~\ref{transfer}, by considering MSE as $\delta (\cdot,\cdot) $ in Eq.~(\ref{TransfAbility}). It can be observed that L-MVAE converges quickly when learning the probabilistic representation of a new database. The baseline is considered as our model trained on a single dataset, MNIST. The average reconstruction errors, calculated using Eq.~(\ref{TransfAbility}) are provided in Figures~\ref{transfer2} a-c for Fashion, SVHN and CIFAR10 databases. The proposed approach adapts quickly to learning a new task when compared to the baseline. We further investigate the difference of the knowledge transfer ability when learning the tasks in a different order. We train our model under the CelebA to CACD and CelebA to Omniglot, respectively. Then we measure the negative log-likelihood of the model for the second task and the results are presented in Fig.~\ref{transfer3}. It can be observed that learning CACD as the prior task can significantly accelerate the convergence when the future task shares similar visual concepts to the prior task. 

\subsection{Studying the over-regularization factors during training}

In this section, we discuss the over-regularization problem in the proposed L-MVAE mixture model. A strong penalty on the KL divergence term in the VAE framework \cite{VAE} can allow the variational distribution to match the prior distributions exactly, so $D_{KL}(q({\bf z|x})||p({\bf z})) = 0$. However, this may lead to a poor representation of the underlying data structure for $q({\bf z}|{\bf x}) = p({\bf z})$. To solve this problem, we implement each expert by using $\beta$-VAE \cite{baeVAE}, which includes a penalty term $\beta^\star$ on KL divergence, expressed as~:
\begin{equation}
\begin{aligned}
 \log p ({\bf x}) &\ge  {\mathbb{E}_{{\bf z} \sim {q_{\varepsilon} } ({\bf z}|{\bf x})}[ \log p_{\theta} ({\bf x}|{\bf z})}] - \beta^\star D_{KL} [ q_{\varepsilon} ({\bf z}|{\bf x} )||p ({\bf z}) ].
\end{aligned}
\label{betavae_DefineL_VAE}
\end{equation}

In the beginning of the training, we use a small $\beta^\star$ which gradually increases up to 1.0, during each task training, in the mixture objective function $L_{L-MVAE}$ from Eq.~(\ref{MELBO}), after replacing $\mathcal{L}^i_{VAE}({\bf x})$ by using Eq.~\eqref{betavae_DefineL_VAE}. We train the mixture model L-MVAE under the MNIST, Fashion, SVHN and CIFAR10 lifelong setting (MFSC sequence) as well as when considering learning these databases in reversed order, denoted as CSFM. We evaluate the Inception Score (IS) on 5000 testing samples from CIFAR10 and the corresponding reconstructions obtained by L-MVAE-MFSC and L-MVAE-CSFM, representing when training with``MFSC'' and ``CSFM'', respectively. The reconstruction results measured by Mean Squared Error (MSE), the structural similarity index measure (SSIM) \cite{Reconstruction_criteria} and Peak-Signal-to-Noise Ratio (PSNR) \cite{Reconstruction_criteria}, provided in Table~\ref{Tab5}, show that L-MVAE achieves competitive results when compared to BatchEnsemble \cite{BatchEnsemble}, a state of the art ensemble model, trained only on CIFAR10. The results also show that the order of learning the four databases does not have a significant impact on the L-MVAE training.

\subsection{The results for the expandable mixture model}

In this section, we evaluate the performance of the proposed expansion mechanism and compare to other generative models, such as BatchEnsemble \cite{BatchEnsemble}. In order to allow BatchEnsemble to perform unsupervised learning tasks, we implement each ensemble member as a VAE. We use MSE, SSIM and PSNR for the evaluation of image reconstruction quality. We train L-MVAE and BatchEnsemble under MNIST, Fashion, SVHN and CIFAR10 lifelong learning. We consider $S^\star = 600$ and L-MVAE adds three new components in the mixture model following the training. The performance of the reconstruction is provided in Table~\ref{Unsupervised2}, where L-MVAE-Dynamic outperforms BatchEnsemble on three criteria. 
The classification tasks when for the lifelong learning of MNIST, Fashion, SVHN, and CIFAR10  are provided in Table~\ref{acc_table}. After the training, L-MVAE-Dynamic has four components and outperforms BatchEnsemble. We also consider a long sequence of tasks~: MNIST, SVHN, Fashion, InverseFashion (IFashion), InverseMNIST (IMNIST), RatedFashion (RFashion), CIFAR10 (MSFIIRC), where we consider $S^\star = 400$ in Eq.~(\ref{AddNewComponent}) for MSFIIRC and provide the results in Tabel~\ref{long_acc_table} where L-MVAE-Dynamic has five components after the lifelong learning. The first and the third components are reused when learning RMNIST and RFashion, respectively, which demonstrates that the appropriate expert that shares similar knowledge with a future task is chosen. Also, L-MVAE-Dynamic achieves the best results in each task when compared to BatchEnsemble.

\section{Conclusion}
\label{Con}

This paper proposes a novel mixture system able to learn successively several tasks, called Lifelong Mixtures of VAEs (L-MVAE) model. Each time when a new database is available, L-MVAE model adapts its weights in order to learn its corresponding probabilistic representation, without forgetting the information learnt from the previous tasks. A mixing-coefficient is used to determine which experts are activated or inactivated during the lifelong learning, preventing catastrophic forgetting. The L-MVAE model is also enabled with an expanding component mechanism, which depends on the complexity of the new task with respect to those already learnt. The proposed lifelong learning framework is applied for supervised, unsupervised and in semi-supervised learning.

{\small
	\bibliographystyle{IEEEtran}
	\bibliography{VAEGAN}

\begin{thebibliography}{10}
\providecommand{\url}[1]{#1}
\csname url@samestyle\endcsname
\providecommand{\newblock}{\relax}
\providecommand{\bibinfo}[2]{#2}
\providecommand{\BIBentrySTDinterwordspacing}{\spaceskip=0pt\relax}
\providecommand{\BIBentryALTinterwordstretchfactor}{4}
\providecommand{\BIBentryALTinterwordspacing}{\spaceskip=\fontdimen2\font plus
\BIBentryALTinterwordstretchfactor\fontdimen3\font minus
  \fontdimen4\font\relax}
\providecommand{\BIBforeignlanguage}[2]{{%
\expandafter\ifx\csname l@#1\endcsname\relax
\typeout{** WARNING: IEEEtran.bst: No hyphenation pattern has been}%
\typeout{** loaded for the language `#1'. Using the pattern for}%
\typeout{** the default language instead.}%
\else
\language=\csname l@#1\endcsname
\fi
#2}}
\providecommand{\BIBdecl}{\relax}
\BIBdecl

\bibitem{LifeLong_review}
G.~Parisi, R.~Kemker, J.~Part, C.~Kanan, and S.~Wermter, ``Continual lifelong
  learning with neural networks: A review,'' in \emph{Proc. of the ACM India
  Joint Int. Conf. on Data Science and Management of Data}, 2019, pp. 362--365.

\bibitem{Adanet}
C.~Cortes, X.~Gonzalvo, V.~Kuznetsov, M.~Mohri, and S.~Yang, ``{AdaNet}:
  Adaptive structural learning of artificial neural networks,'' in \emph{Proc.
  of Int. Conf. on Machine Learning (ICML), vol. PMLR 70}, 2017, pp. 874--883.

\bibitem{ProgressiveNN}
\BIBentryALTinterwordspacing
A.~Rusu, N.~Rabinowitz, G.~Desjardins, H.~Soyer, J.~Kirkpatrick,
  K.~Kavukcuoglu, R.~Pascanu, and R.~Hadsell, ``Progressive neural networks,''
  2016. [Online]. Available: \url{https://arxiv.org/abs/1606.04671}
\BIBentrySTDinterwordspacing

\bibitem{Error_driven}
T.~Xiao, J.~Zhang, K.~Yang, Y.~Peng, and Z.~Zhang, ``Error-driven incremental
  learning in deep convolutional neural network for large-scale image
  classification,'' in \emph{Proc. of ACM Int. Conf. on Multimedia}, 2014, pp.
  177--186.

\bibitem{OnlineLearning}
G.~Zhou, K.~Sohn, and H.~Lee, ``Online incremental feature learning with
  denoising autoencoders,'' in \emph{Proc. Int. Conf. on Artificial
  Intelligence and Statistics (AISTATS), vol. PMLR 22}, 2012, pp. 1453--1461.

\bibitem{LearnAdd}
R.~Polikar, L.~Upda, S.~Upda, and V.~Honavar, ``Learn++: An incremental
  learning algorithm for supervised neural networks,'' \emph{IEEE Trans. on
  Systems Man and Cybernetics, Part C}, vol.~31, no.~4, pp. 497--508, 2001.

\bibitem{Distilling_nets}
\BIBentryALTinterwordspacing
G.~Hinton, O.~Vinyals, and J.~Dean, ``Distilling the knowledge in a neural
  network,'' in \emph{NIPS Deep Learning and Representation Learning Workshop},
  2015. [Online]. Available: \url{http://arxiv.org/abs/1503.02531}
\BIBentrySTDinterwordspacing

\bibitem{LessForgetting}
H.~Jung, J.~Ju, M.~Jung, and J.~Kim, ``Less-forgetting learning in deep neural
  networks,'' in \emph{Proc. AAAI Conf. on Artif. Intel.}, 2016, pp.
  3358--3365.

\bibitem{EWC}
J.~Kirkpatrick, R.~Pascanu, N.~Rabinowitz, J.~Veness, G.~Desjardins, A.~Rusu,
  K.~Milan, J.~Quan, T.~Ramalho, A.~Grabska-Barwinska, D.~Hassabis, C.~Clopath,
  D.~Kumaran, and R.~Hadsell, ``Overcoming catastrophic forgetting in neural
  networks,'' \emph{Proc. of the National Academy of Sciences (PNAS)}, vol.
  114, no.~13, pp. 3521--3526, 2017.

\bibitem{Lwf}
Z.~Li and D.~Hoiem, ``Learning without forgetting,'' \emph{IEEE Trans. on
  Pattern Analysis and Machine Intelligence}, vol.~40, no.~12, pp. 2935--2947,
  2017.

\bibitem{LifeLong_combination}
B.~Ren, H.~Wang, J.~Li, and H.~Gao, ``Life-long learning based on dynamic
  combination model,'' \emph{Applied Soft Computing}, vol.~56, pp. 398--404,
  2017.

\bibitem{TinyLifelong}
\BIBentryALTinterwordspacing
A.~Chaudhry, M.~Rohrbach, M.~Elhoseiny, T.~Ajanthan, P.~Dokania, P.~H.~S. Torr,
  and M.~Ranzato, ``On tiny episodic memories in continual learning,'' in
  \emph{Proc. ICML Workshop on Multi-Task and Lifelong Reinforcement Learning},
  2019. [Online]. Available: \url{https://arxiv.org/abs/1902.10486}
\BIBentrySTDinterwordspacing

\bibitem{GradientLifelong}
R.~Aljundi, M.~Lin, B.~Goujaud, and Y.~Bengio, ``Gradient based sample
  selection for online continual learning,'' in \emph{Advances in Neural Inf.
  Proc. Systems (NeurIPS)}, 2019, pp. 11\,816--11\,825.

\bibitem{Generative_replay}
H.~Shin, J.~K. Lee, J.~Kim, and J.~Kim, ``Continual learning with deep
  generative replay,'' in \emph{Advances in Neural Inf. Proc. Systems
  (NeurIPS)}, 2017, pp. 2990--2999.

\bibitem{GenerativeLifelong}
\BIBentryALTinterwordspacing
J.~Ramapuram, M.~Gregorova, and A.~Kalousis, ``Lifelong generative modeling,''
  in \emph{Proc. Int. Conf. on Learning Representations (ICLR)}, 2018.
  [Online]. Available: \url{https://arxiv.org/pdf/1705.09847.pdf}
\BIBentrySTDinterwordspacing

\bibitem{Lifelong_VAE}
A.~Achille, T.~Eccles, L.~Matthey, C.~Burgess, N.~Watters, A.~Lerchner, and
  I.~Higgins, ``Life-long disentangled representation learning with
  cross-domain latent homologies,'' in \emph{Advances in Neural Inf. Proc.
  Systems (NeurIPS)}, 2018, pp. 9873--9883.

\bibitem{LifelongUnsupervisedVAE}
\BIBentryALTinterwordspacing
D.~Rao, F.~Visin, A.~Rusu, R.~Teh, Y. W.~Pascanu, and R.~Hadsell, ``Continual
  unsupervised representation learning,'' in \emph{Advances in Neural Inf.
  Proc. Systems (NeurIPS)}, 2019. [Online]. Available:
  \url{https://arxiv.org/abs/1910.14481}
\BIBentrySTDinterwordspacing

\bibitem{LGAN}
M.~Zhai, L.~Chen, F.~Tung, J.~He, M.~Nawhal, and G.~Mori, ``Lifelong {GAN}:
  Continual learning for conditional image generation,'' in \emph{Proc. IEEE
  Int. Conf. on Computer Vision (ICCV)}, 2019, pp. 2759--2768.

\bibitem{MemoryGAN}
C.~Wu, L.~Herranz, X.~Liu, J.~van~de Weijer, and B.~Raducanu, ``Memory replay
  gans: Learning to generate new categories without forgetting,'' in
  \emph{Advances In Neural Inf. Proc. Systems (NeurIPS)}, 2018, pp. 5962--5972.

\bibitem{ImprovedAGEM}
\BIBentryALTinterwordspacing
Y.~Guo, M.~Liu, T.~Yang, and T.~Rosing, ``Improved schemes for episodic
  memory-based lifelong learning,'' in \emph{Advances in Neural Information
  Processing Systems (NeurIPS)}, 2020. [Online]. Available:
  \url{https://arxiv.org/abs/1909.11763}
\BIBentrySTDinterwordspacing

\bibitem{CalibratingCNN}
P.~Singh, V.~K. Verma, P.~Mazumder, L.~Carin, and P.~Rai, ``Calibrating {CNNs}
  for lifelong learning,'' in \emph{Advances in Neural Information Processing
  Systems (NeurIPS)}, 2020.

\bibitem{ThreeScenarios_Lifelong}
\BIBentryALTinterwordspacing
G.~van~de Ven and A.~Tolias, ``Three scenarios for continual learning,'' in
  \emph{NeurIPS Continual Learning workshop}, 2018. [Online]. Available:
  \url{https://arxiv.org/abs/1904.07734}
\BIBentrySTDinterwordspacing

\bibitem{Continual_Learning}
F.~Zenke, B.~Poole, and S.~Ganguli, ``Continual learning through synaptic
  intelligence,'' in \emph{Proc. of Int. Conf. on Machine Learning (ICML), vol.
  PMLR 70}, 2017, pp. 3987--3995.

\bibitem{VMVAE}
Y.~Shi, N.~Siddharth, B.~Paige, and P.~Torr, ``Variational mixture-of-experts
  autoencoders for multi-modal deep generative models,'' in \emph{Advances in
  Neural Inf. Proc. Systems (NeurIPS)}, 2019, pp. 15\,718--15\,729.

\bibitem{MSVAE}
R.~Kurle, S.~Günnemann, and P.~van~der Smagt, ``Multi-source neural
  variational inference,'' in \emph{Proc. AAAI Conf. on Artificial
  Intelligence}, 2019, pp. 4114--4121.

\bibitem{Cluster_VAE}
\BIBentryALTinterwordspacing
N.~Dilokthanakul, P.~Mediano, M.~Garnelo, M.~Lee, H.~Salimbeni, K.~Arulkumaran,
  and M.~Shanahan, ``Deep unsupervised clustering with {G}aussian mixture
  variational autoencoders,'' in \emph{Proc. Int. Conf. on Learning
  Representations (ICLR)}, 2018. [Online]. Available:
  \url{https://arxiv.org/abs/1611.02648}
\BIBentrySTDinterwordspacing

\bibitem{Multimodal_generative}
M.~Wu and N.~Goodman, ``Multimodal generative models for scalable
  weakly-supervised learning,'' in \emph{Advances in Neural Information
  Processing Systems}, 2018, pp. 5575--5585.

\bibitem{VAE}
\BIBentryALTinterwordspacing
D.~P. Kingma and M.~Welling, ``Auto-encoding variational {B}ayes,'' 2013.
  [Online]. Available: \url{https://arxiv.org/abs/1312.6114}
\BIBentrySTDinterwordspacing

\bibitem{baeVAE}
I.~Higgins, L.~Matthey, A.~Pal, C.~Burgess, X.~Glorot, M.~Botvinick,
  S.~Mohamed, and A.~Lerchner, ``$\beta$-{VAE}: Learning basic visual concepts
  with a constrained variational framework,'' in \emph{Proc. Int. Conf. on
  Learning Representations (ICLR)}, 2017.

\bibitem{DisentanglingByFactorising}
H.~Kim and A.~Mnih, ``Learning disentangled joint continuous and discrete
  representations,'' in \emph{Proc. Int. Conf. on Machine Learning (ICML), PMLR
  80}, 2018, pp. 2649--2658.

\bibitem{VAETCE}
S.~Gao, R.~Brekelmans, G.~V. Steeg, and A.~Galstyan, ``Auto-encoding total
  correlation explanation,'' in \emph{Proc. Int. Conf. on Art. Intel. and Stat.
  (AISTATS), vol. PMLR 89}, 2019, pp. 1157--1166.

\bibitem{UnVAE}
\BIBentryALTinterwordspacing
C.~P. Burgess, I.~Higgins, A.~Pal, L.~Matthey, N.~Watters, G.~Desjardins, and
  A.~Lerchner, ``Understanding disentangling in $\beta$-{VAE},'' in \emph{NIPS
  Workshop on Learning Disentangled Representations}, 2018. [Online].
  Available: \url{https://arxiv.org/abs/1804.03599}
\BIBentrySTDinterwordspacing

\bibitem{DisentanglingDistanglement}
E.~Mathieu, T.~Rainforth, N.~Siddharth, and Y.~W. Teh, ``Disentangling
  disentanglement in variational autoencoders,'' in \emph{Proc. Int. Conf. on
  Machine Learning (ICML), PMLR 97}, 2019, pp. 4402--4412.

\bibitem{Catastrophic}
R.~M. French, ``Catastrophic forgetting in connectionist networks,''
  \emph{Trends in cognitive sciences}, vol.~3, no.~4, pp. 128--135, 1999.

\bibitem{OnlineStructuredLaplace}
H.~Ritter, A.~Botev, and D.~Barber, ``Online structured {L}aplace
  approximations for overcoming catastrophic forgetting,'' in \emph{Advances in
  Neural Information Processing Systems (NeurIPS)}, vol.~31, 2018, pp.
  3742--3752.

\bibitem{LTS}
F.~Ye and A.~G. Bors, ``Lifelong {T}eacher-{S}tudent network learning,''
  \emph{IEEE Trans. on Pattern Analysis and Machine Intelligence}, 2021.

\bibitem{LatentVAEGAN}
------, ``Learning latent representations across multiple data domains using
  lifelong {VAEGAN},'' in \emph{Proc. European Conf. on Computer Vision (ECCV),
  vol LNCS 12365}, 2020, pp. 777--795.

\bibitem{GAN}
I.~Goodfellow, J.~Pouget-Abadie, M.~Mirza, B.~Xu, D.~Warde-Farley, S.~Ozair,
  A.~Courville, and Y.~Bengio, ``Generative adversarial nets,'' in
  \emph{Advances in Neural Inf. Proc. Systems (NIPS)}, 2014, pp. 2672--2680.

\bibitem{Expertgate}
R.~Aljundi, P.~Chakravarty, and T.~Tuytelaars, ``Expert gate: Lifelong learning
  with a network of experts,'' in \emph{Proc. of IEEE Conf. on Computer Vision
  and Pattern Recognition (CVPR)}, 2017, pp. 3366--3375.

\bibitem{BoostingTransfer}
W.~Dai, Q.~Yang, G.~R. Xue, and Y.~Yu, ``Boosting for transfer learning,'' in
  \emph{Proc. Int. Conf. on Machine Learning (ICML)}, 2007, pp. 193--200.

\bibitem{VCL}
\BIBentryALTinterwordspacing
C.~V. Nguyen, Y.~Li, T.~D. Bui, and R.~E. Turner, ``Variational continual
  learning,'' in \emph{Proc. Int. Conf. on Learning Representations (ICLR)},
  2018. [Online]. Available: \url{https://arxiv.org/abs/1710.10628}
\BIBentrySTDinterwordspacing

\bibitem{CL_Bayesian}
R.~Kurle, B.~Cseke, A.~Klushyn, P.~van~der Smagt, and S.~Günnemann,
  ``Continual learning with bayesian neural networks for non-stationary data,''
  in \emph{Proc. Int. Conf. on Learning Representations (ICLR)}, 2020.

\bibitem{AGEM}
\BIBentryALTinterwordspacing
A.~Chaudhry, M.~Ranzato, M.~Rohrbach, and M.~Elhoseiny, ``Efficient lifelong
  learning with {A-GEM},'' in \emph{Proc. Int. Conf. on Learning
  Representations (ICLR)}, 2019. [Online]. Available:
  \url{https://arxiv.org/abs/1812.00420}
\BIBentrySTDinterwordspacing

\bibitem{GradientEpisodic}
D.~Lopez-Paz and M.~Ranzato, ``Gradient episodic memory for continual
  learning,'' in \emph{Advances in Neural Information Processing Systems},
  2017, pp. 6467--6476.

\bibitem{MixVAE}
F.~Ye and A.~G. Bors, ``Deep mixture generative autoencoders,'' \emph{IEEE
  Trans. on Neural Networks and Learning Systems}, 2021.

\bibitem{VGMM}
N.~Nasios and A.~G. Bors, ``Variational learning for {G}aussian mixtures,''
  \emph{IEEE Trans. on Systems, Man, and Cybernetics, Part B (Cybernetics)},
  vol.~36, no.~4, pp. 849--862, 2006.

\bibitem{SparseMulGMR}
L.~Weruaga and J.~Via, ``Sparse multivariate {G}aussian mixture regression,''
  \emph{IEEE Trans. on Neural Networks and Learning Systems}, vol.~26, no.~5,
  pp. 1098--1108, 2015.

\bibitem{Gumble_softmax}
\BIBentryALTinterwordspacing
E.~Jang, S.~Gu, and B.~Poole, ``Categorical reparameterization with
  {Gumbel-Softmax},'' in \emph{Proc. Int. Conf. on Learning Representations
  (ICLR)}, 2017. [Online]. Available: \url{https://arxiv.org/abs/1611.01144}
\BIBentrySTDinterwordspacing

\bibitem{GumbelMaxTrick}
E.~J. Gumbel, ``Statistical theory of extreme values and some practical
  applications,'' \emph{NBS Applied Mathematics Series}, vol.~33, 1954.

\bibitem{Semi_VAE}
S.~Narayanaswamy, B.~Paige, J.-W. Van~de Meent, A.~Desmaison, N.~Goodman,
  P.~Kohli, F.~Wood, and P.~Torr, ``Learning disentangled representations with
  semi-supervised deep generative models,'' in \emph{Advances in Neural Inf.
  Proc. Systems (NeurIPS)}, 2017, pp. 5925--5935.

\bibitem{MNIST}
Y.~LeCun, L.~Bottou, Y.~Bengio, and P.~Haffner, ``Gradient-based learning
  applied to document recognition,'' \emph{Proc. of the IEEE}, vol.~86, no.~11,
  pp. 2278--2324, 1998.

\bibitem{FashionMNIST}
\BIBentryALTinterwordspacing
H.~Xiao, K.~Rasul, and R.~Vollgraf, ``Fashion-{MNIST}: a novel image dataset
  for benchmarking machine learning algorithms,'' 2017. [Online]. Available:
  \url{https://arxiv.org/abs/1708.07747}
\BIBentrySTDinterwordspacing

\bibitem{SVHN}
Y.~Netzer, T.~Wang, A.~Coates, A.~Bissacco, B.~Wu, and A.~Y. Ng, ``Reading
  digits in natural images with unsupervised feature learning,'' in \emph{NIPS
  Workshop on Deep Learning and Unsupervised Feature Learning}, 2011.

\bibitem{CIFAR10}
A.~Krizhevsky and G.~Hinton, ``Learning multiple layers of features from tiny
  images,'' Tech. Rep., 2009.

\bibitem{SemiSuper}
D.~P. Kingma, S.~Mohamed, D.~J. Rezende, and M.~Welling, ``Semi-supervised
  learning with deep generative models,'' in \emph{Advances in Neural Inf.
  Proc. Systems (NIPS)}, 2014, pp. 3581--3589.

\bibitem{InceptScore}
T.~Salimans, I.~Goodfellow, W.~Zaremba, V.~Cheung, A.~Radford, and X.~Chen,
  ``Improved techniques for training {GANs},'' in \emph{Advances in Neural Inf.
  Proc. Systems (NIPS)}, 2016, pp. 2234--2242.

\bibitem{MixWass}
\BIBentryALTinterwordspacing
S.~Arora, R.~Ge, Y.~Liang, T.~Ma, and Y.~Zhang, ``Generalization and
  equilibrium in generative adversarial nets ({GAN}s),'' in \emph{Proc. Int.
  Conf. on Machine Learning (ICML), vol. PMLR 70}, 2017, pp. 224--232.
  [Online]. Available: \url{arXiv preprint arXiv:1703.00573}
\BIBentrySTDinterwordspacing

\bibitem{ConvGAN}
\BIBentryALTinterwordspacing
A.~Radford, L.~Metz, and S.~Chintala, ``Unsupervised representation learning
  with deep convolutional generative adversarial networks,'' in \emph{Proc.
  Int. Conf. on Learning Representations (ICLR)}, 2016. [Online]. Available:
  \url{arXiv preprint arXiv:1511.06434}
\BIBentrySTDinterwordspacing

\bibitem{VAE_symmetric}
\BIBentryALTinterwordspacing
L.~Chen, S.~Dai, Y.~Pu, C.~Li, Q.~Su, and L.~Carin, ``Symmetric variational
  autoencoder and connections to adversarial learning,'' 2017. [Online].
  Available: \url{https://arxiv.org/abs/1709.01846}
\BIBentrySTDinterwordspacing

\bibitem{AdLearnInf}
\BIBentryALTinterwordspacing
V.~Dumoulin, I.~Belghazi, B.~Poole, O.~Mastropietro, A.~Lamb, M.~Arjovsky, and
  A.~Courville, ``Adversarially learned inference,'' in \emph{Proc. Int. Conf.
  on Learning Representations (ICLR)}, 2017. [Online]. Available: \url{arXiv
  preprint arXiv:1606.00704}
\BIBentrySTDinterwordspacing

\bibitem{Pixelcnn++}
\BIBentryALTinterwordspacing
T.~Salimans, A.~Karpathy, X.~Chen, and D.~P. Kingma, ``{PixelCNN++}: Improving
  the {PixelCNN} with discretized logistic mixture likelihood and other
  modifications,'' in \emph{Proc. Int. Conf. on Learning Representations
  (ICLR)}, 2017. [Online]. Available: \url{https://arxiv.org/abs/1701.05517}
\BIBentrySTDinterwordspacing

\bibitem{ASymVAE}
Y.~Pu, W.~Wang, R.~Henao, C.~L., Z.~Gan, C.~Li, and L.~Carin, ``Adversarial
  symmetric variational autoencoder,'' in \emph{Advances in Neural Inf. Proc.
  Systems (NeurIPS)}, 2017, pp. 4333--4342.

\bibitem{BatchEnsemble}
\BIBentryALTinterwordspacing
Y.~Wen, D.~Tran, and J.~Ba, ``{BatchEnsemble}: an alternative approach to
  efficient ensemble and lifelong learning,'' in \emph{Proc. Int. Conf. on
  Learning Representations (ICLR)}, 2020. [Online]. Available:
  \url{https://arxiv.org/abs/2002.06715}
\BIBentrySTDinterwordspacing

\bibitem{Reconstruction_criteria}
A.~Hore and D.~Ziou, ``Image quality metrics: {PSNR} vs. {SSIM},'' in
  \emph{Proc. Int. Conf. on Pattern Recognition (ICPR)}, 2010, pp. 2366--2369.

\end{thebibliography}
}

\begin{IEEEbiography}[{\includegraphics[width=1in,height=1.25in,clip,keepaspectratio]{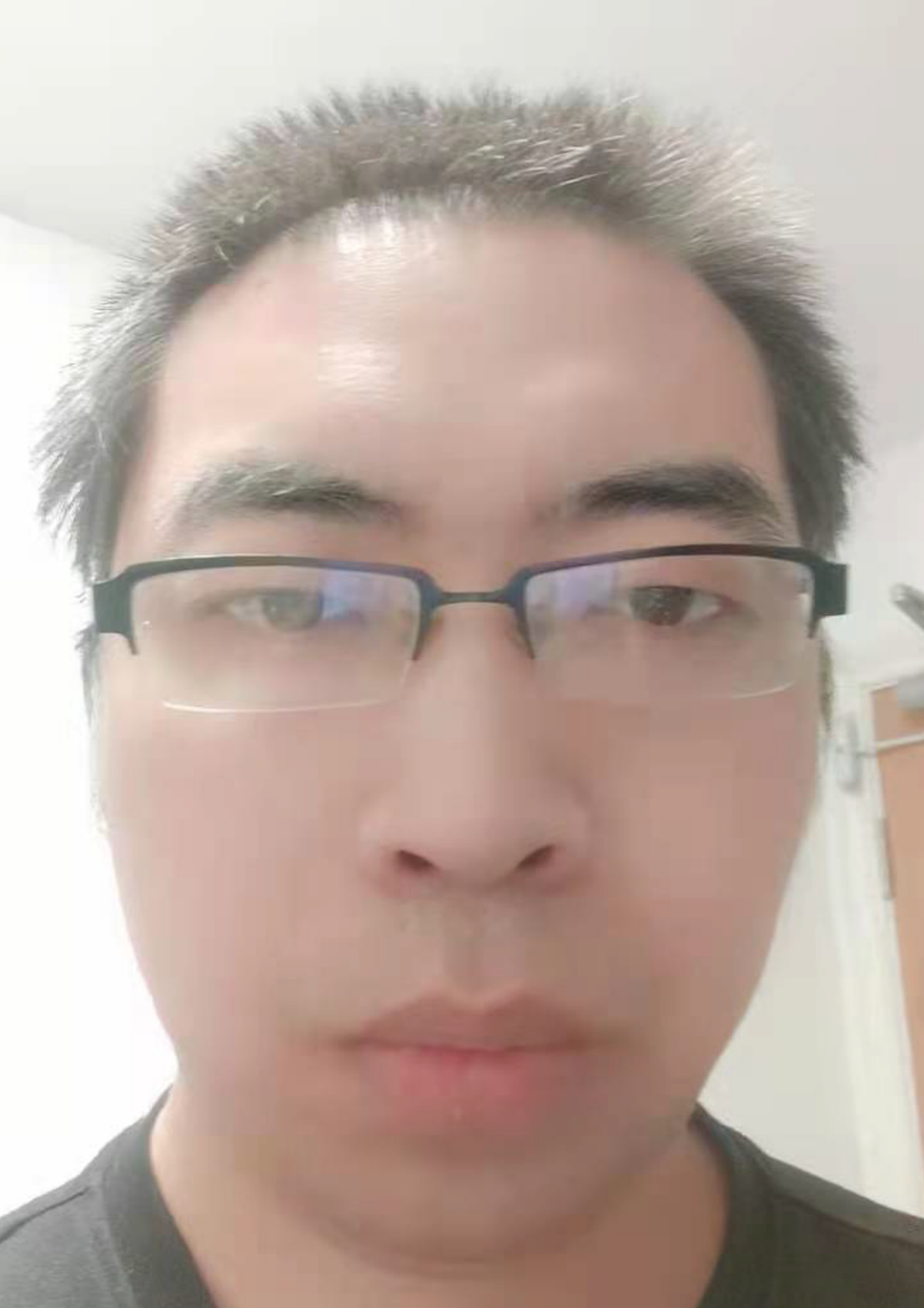}}]{Fei Ye} is a currently third-year PHD student in computer science from University of York. He received the bachelor degree from Chengdu University of Technology, China, in 2014 and the master degree in computer science and technology from Southwest Jiaotong University, China, in 2018. His research topic includes deep generative image model, lifelong learning and mixture models.
\end{IEEEbiography}

\begin{IEEEbiography}[{\includegraphics[width=1in,height=1.25in,clip,keepaspectratio]{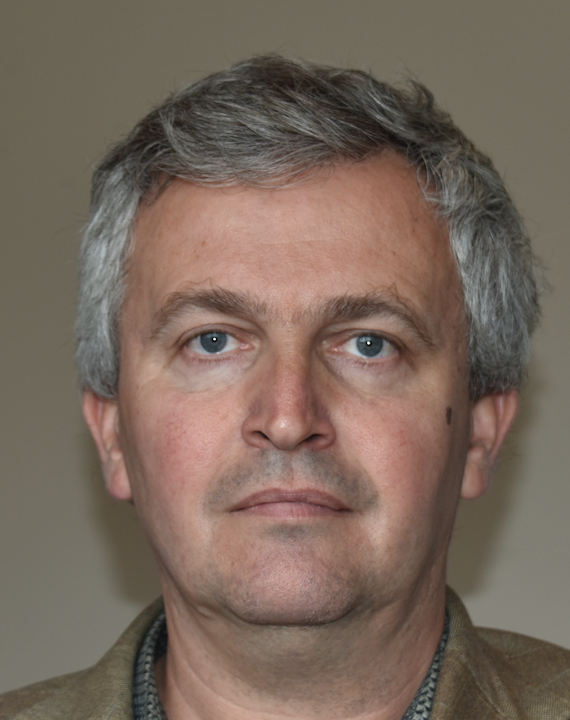}}]{Adrian G. Bors} (Senior Member, IEEE)
received the M.Sc. degree in electronics engineering from the Polytechnic University of Bucharest, Bucharest, Romania, in 1992, and the Ph.D. degree in informatics from the University of Thessaloniki, Thessaloniki, Greece, in 1999.
In 1999, he joined the Department of Computer
Science, University of York, York, U.K., where he
is currently a Lecturer.

In 1999 he joined the Department of Computer Science, Univ. of York, U.K., where he is currently a lecturer. Dr. Bors was a Research Scientist at Tampere Univ. of Technology, Finland, a Visiting Scholar at the Univ. of California at San Diego (UCSD), and an Invited Professor at the Univ. of Montpellier, France. 
Dr. Bors has authored and co-authored more than 140 research papers including 32 in journals. His research interests include computer vision, computational intelligence and image processing.

Dr. Bors has been a member of the organizing committees for IEEE WIFS 2021, IPTA 2020, IEEE ICIP 2018, BMVC 2016, IPTA 2014, CAIP 2013, and IEEE
ICIP 2001. He was an Associate Editor of the IEEE TRANSACTIONS ON IMAGE PROCESSING from 2010 to 2014 and the IEEE TRANSACTIONS ON NEURAL NETWORKS from 2001 to 2009. He was a Co-Guest Editor for a
special issue on Machine Vision for the International Journal for Computer Vision in 2018 and the Journal of Pattern Recognition in 2015.
\end{IEEEbiography}

\end{document}